\documentclass[twoside,11pt]{article}
\usepackage{jmlr2e}
\usepackage{amsmath}
\usepackage{array}
\usepackage{multirow}
\usepackage{hhline}
\usepackage{pgfplots}
\pgfplotsset{compat=1.14}
\usepackage{subcaption}
\usepackage{hyperref}
\usepackage[ruled]{algorithm2e}
\usepackage{algorithmicx}
\usepackage{booktabs}
\usepackage{setspace}

\ShortHeadings{Explaining Data-Driven Decisions}{Fern\'{a}ndez-Lor\'{i}a, Provost and Han}
\firstpageno{1}

\author{\name Carlos Fern\'{a}ndez-Lor\'{i}a \email cfernand@stern.nyu.edu \\
       \addr New York University
       \AND
       \name Foster Provost \email fprovost@stern.nyu.edu \\
       \addr New York University
       \AND
       \name Xintian Han \email xintian.han@nyu.edu \\
       \addr New York University}

\RequirePackage{natbib}

\editor{}

\title{Explaining Data-Driven Decisions made by AI Systems: \\ The Counterfactual Approach}

\begin{document}

\maketitle

\begin{abstract} 
We examine counterfactual explanations for explaining the decisions made by model-based AI systems. The counterfactual approach we consider defines an explanation as a set of the system's data inputs that causally drives the decision (i.e., changing the inputs in the set changes the decision) and is irreducible (i.e., changing any subset of the inputs does not change the decision).
We (1) demonstrate how this framework may be used to provide explanations for decisions made by general data-driven AI systems that can incorporate features with arbitrary data types and multiple predictive models, and (2) propose a heuristic procedure to find the most useful explanations depending on the context. We then contrast counterfactual explanations with methods that explain model predictions by weighting features according to their importance (e.g., Shapley additive explanations (SHAP), local interpretable model-agnostic explanations (LIME))  and present two fundamental reasons why we should carefully consider whether importance-weight explanations are well suited to explain system decisions. Specifically, we show that (i) features with a large importance weight for a model prediction may not affect the corresponding decision, and (ii) importance weights are insufficient to communicate whether and how features influence decisions. We demonstrate this with several concise examples and three detailed case studies that compare the counterfactual approach with SHAP to illustrate conditions under which counterfactual explanations explain data-driven decisions better than importance weights.

\end{abstract}


\clearpage

\maketitle

\section{Introduction}
Artificial intelligence (AI) systems use data and predictive models to make decisions across many applications and industries. In many cases, the ability to explain system decisions is critical for the success of the system. For example, explanations may help customers understand the reasoning behind decisions that affect them. Managers and analysts may use explanations to learn about the domain in which the system is being used. Data scientists and machine learning engineers may also use the explanations to identify, debug, and address potential flaws in the system. 

Stakeholders can also be skeptical and reluctant to adopt AI systems without the ability to explain system decisions, even if the systems have been shown to improve decision-making performance~\citep{arnold2006differential, kayande2009incorporating}. In fact, many data-rich organizations struggle when adopting AI decision-making systems because of managerial and cultural challenges rather than issues related to data and technology~\citep{lavalle2011big}. Thus, many researchers have tried to reduce the gap in stakeholders’ understanding of AI systems by proposing methods for explaining predictive models and their predictions. 

Methods for explaining AI models and their predictions include extracting rules that represent the inner workings of the model \citep[e.g.,][]{craven1996extracting,jacobsson2005rule,martens2007comprehensible} and associating weights to each feature according to their importance for model predictions \citep[e.g., ][]{lundberg2017unified, ribeiro2016should}. Importance weights have become increasingly popular because of advances in ``model-agnostic'' methods that can produce importance weights for any predictive model: the weights explain predictions in terms of features, so users can understand any specific prediction without any knowledge of the underlying model or the modeling method(s). For example, two popular methods for explaining model predictions, local interpretable model-agnostic explanations (LIME)~\citep{ribeiro2016should} and Shapley additive explanations (SHAP)~\citep{lundberg2017unified}, are model-agnostic and produce importance-weight explanations.

This paper contributes to the AI-explanation literature in the following ways:
\begin{enumerate}
    \item We show that explaining model predictions and explaining the decisions of a system-in-practice are not the same type of task.
    \item We demonstrate that importance-weight methods are not well suited to explain system decisions despite their popularity.
    \item We propose a generalized framework based on counterfactual reasoning that can produce context-dependent explanations for decisions made by general, data-driven AI systems.
\end{enumerate}

To support our first contribution, we use multiple examples to show that features that have a large impact on a prediction may not necessarily affect the decision made using that prediction. Conversely, features that affect a system decision may not have a substantive impact on the prediction on which the decision was based, when that impact is calculated using the most common feature-importance method. As a result, importance weights designed to explain model predictions may yield an inaccurate picture of how input data affect system decisions. 

Our second contribution shows that identifying and quantifying important features is not sufficient to explain system decisions, even when importance is assessed with respect to the decisions being explained. As an example, suppose that a credit scoring system denies credit to a loan applicant, and that feature importance weights reveal that the two most important features in the credit denial decision were annual income and loan amount. While informative, this ``explanation" does not explain what made the system deny credit. Would changing either income or loan amount be enough for the system to approve credit? Would it be necessary to change both? Perhaps even changing both would not be enough. From the weights alone, it is not clear how the important features may influence the decision. This is not an indictment of methods that calculate feature importance; they were not designed to explain system decisions. However, we are not aware of prior work that clarifies this for research or practice.

Finally, our third contribution proposes a general framework based on counterfactual explanations as an alternative to importance weights. For the question ``why did the model-based system make a specific decision?'', our counterfactual approach asks specifically, ``which data inputs caused the system to make its decision?'' This approach is advantageous because (i) it explains decisions rather than the outputs of the model(s) on which the decisions are based; (ii) it standardizes the form that an explanation can take; (iii) it does not require all features to be part of the explanation, and (iv) the explanations can be separated from the specifics of the model.

Dozens of methods based on counterfactual explanations have been proposed \citep{verma2020counterfactual}, but most explain model predictions rather than system decisions. To our knowledge, the first framework for counterfactual explanations for decisions was introduced in this journal to explain document classifications~\citep{martens2014explaining}. This framework, which is model-agnostic, has also been applied to other sparse high-dimensional settings~\citep{moeyersoms2016explaining,chen2017enhancing,ramon2019counterfactual}, but researchers do not all see how the framework can be generalized to settings beyond text~\citep[see, e.g.,][]{explanationssite,wachter2017counterfactual,biran2017explanation}. Therefore, our third contribution extends the framework introduced by Martens and Provost to provide explanations for decisions made by general, data-driven AI systems that may incorporate features with arbitrary data types and multiple predictive models. In addition, we propose and showcase a heuristic procedure that can search and sort counterfactual explanations according to their context-specific relevance.  

We demonstrate these extensions and illustrate the advantages of the counterfactual approach by comparing it to SHAP~\citep{lundberg2017unified}, an increasingly popular method to explain model predictions that unites several feature importance weighting methods. We present three simple examples showing the advantages, and then we present three business case studies using real-world data to show that the differences between the approaches are not purely academic.

\section{AI Systems and Explanations}

We focus on explaining decisions made by systems that use predictive statistical models to support or automate decision-making~\citep{shmueli2011predictive}, and, in particular, on systems that make or recommend discrete decisions. We refer to these as artificial intelligence (AI) systems. These AI systems may or may not have been built using machine learning; this paper studies explaining the decisions of a system-in-practice, not how the system was built.\footnote{However, explaining the decisions of the system-in-practice can also help to understand the system-building process, for example, by debugging training data~\citep{martens2014explaining}.} 

\subsection{Explaining models and their predictions}

Over the past several decades, many researchers have worked on explaining predictive models, which is not the same as explaining the decisions made with such models, as we discuss in detail later in this paper. Because symbolic models, such as decision trees, are often considered straightforward to explain when they are small, most research has focused on explaining non-symbolic (black box) models or large models.

Rule-based explanations have been a popular approach to explain black-box models. 
For example, in many credit scoring applications, banking regulatory entities require banks to implement globally comprehensible predictive models~\citep{martens2007comprehensible}. Typical techniques to provide rule-based explanations consist of approximating the black-box model with a symbolic model~\citep{craven1996extracting} or extracting explicit if-then rules~\citep{andrews1995survey}. Importantly, these ``global'' explanations attempt to explain the model as a whole rather than explaining particular decisions made. As \cite{martens2014explaining} point out, this can be viewed as explaining every possible decision the model might make, but the methods are not designed to explain individual decisions, which is the focus of our study. Furthermore, the model itself being explainable does not necessarily imply that individual decisions made by the model are explainable.  

Another approach is to produce models explicitly designed to be both accurate and comprehensible~\citep{wang2015falling,angelino2017learning}. However, these studies are about building intelligible systems, not explaining the decisions of a system currently in use. Therefore, those methods are meant to replace existing black-box models with new models that are easier to interpret. This may not always be possible, particularly if the goal is to explain the decisions of a system that incorporates multiple models or subsystems. 

\begin{figure}
    \centering
    \includegraphics[width=1\textwidth]{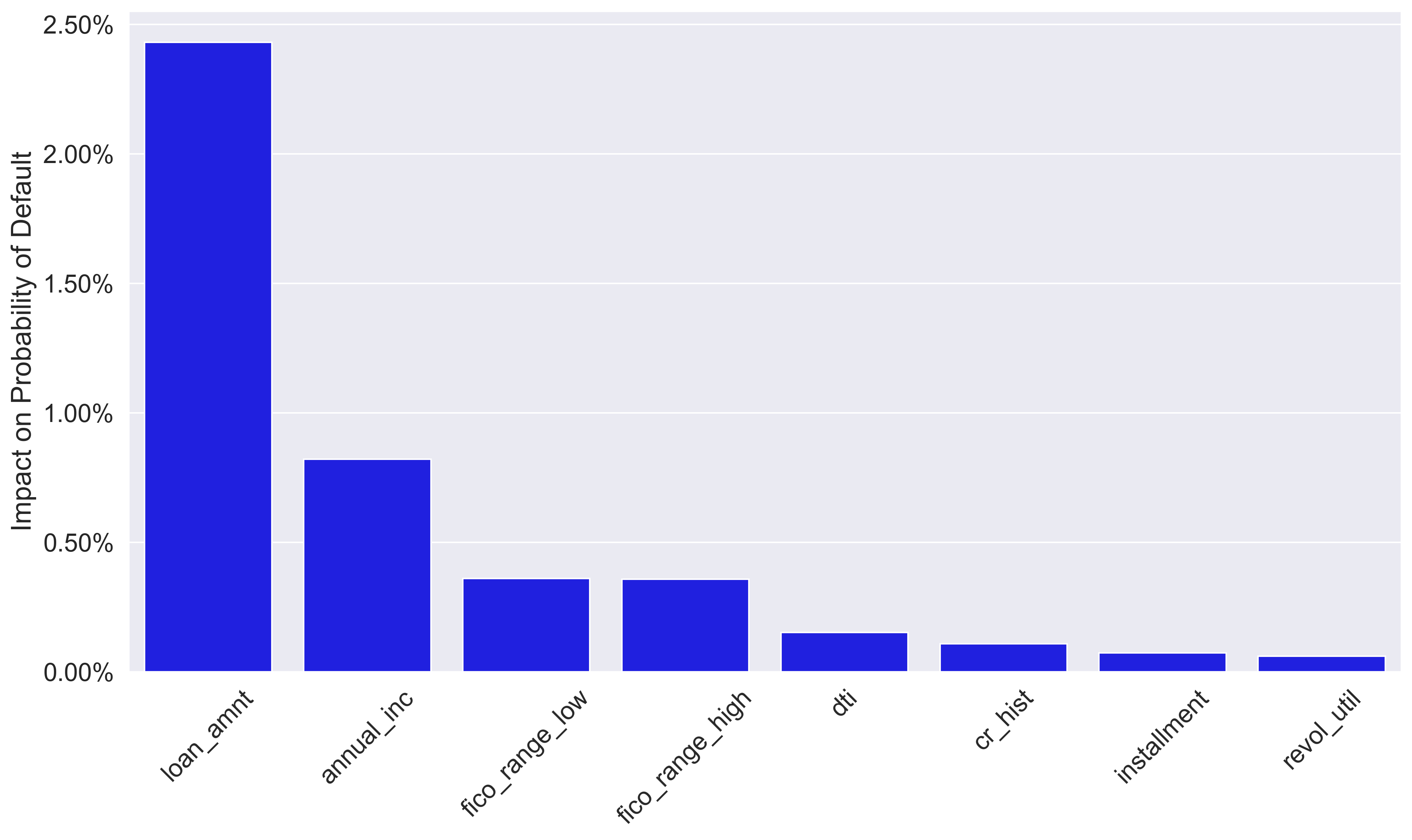}
    \caption{Example of an importance-weight explanation for a model prediction}
    \label{fig:weights_intro}
\end{figure}

A fundamentally different approach (and one of the primary explanation techniques we analyze in this paper) is to explain the predictions of complex models by associating a weight to each feature in the model. Methods that use this approach often decompose each prediction into the individual contributions of each feature and use the decompositions as explanations, allowing one to visualize explanations at the instance level. Continuing with the earlier credit scoring example, Figure~\ref{fig:weights_intro} shows an importance-weight explanation for an individual who has an above-average estimated probability of default (based on one of the case studies we present below). These importance weights were generated using SHAP~\citep{lundberg2017unified}, which we will discuss in the following sections. As the example shows, each weight in the explanation represents the attributed impact of its respective feature on the prediction. Thus, the weight associated with the loan amount feature (`loan\_amnt') implies that the feature is attributed an increase of roughly 2.5\% in the estimated probability of default for that individual.

The main strength of this approach is that the explanations are defined in terms of the domain (i.e., the features), separating them from the specifics of the model being explained. As a result, models can be replaced without replacing the explanation method; end users (such as customers or managers) do not need any knowledge of the underlying modeling methods to understand the explanations, and different models may be compared in terms of their explanations in settings where transparency is critical. These are some reasons why importance-weight methods have become one of the most popular approaches to explain model predictions.

One notable challenge, however, is the computation of the weights. For example, a common way of assessing feature importance is based on simulating lack of knowledge about features~\citep{robnik2008explaining,lemaire2008contact}, typically by comparing the original model’s output with the output obtained when the information given by a specific feature is removed (e.g., by imputing a default value for the feature). Unfortunately, interactions between features may lead to ambiguous explanations because the order in which features are removed may affect the importance attributed to each feature. Researchers have proposed addressing this issue by comparing the model predictions when removing all possible subsets of features~\citep{vstrumbelj2009explaining}, but this is intractable with a large number of features. Therefore, recent formulations (such as SHAP) have attempted to reduce computation time by sampling the space of feature combinations, resulting in sampling-based approximations of the influence of each feature on the prediction~\citep{vstrumbelj2010efficient,ribeiro2016should,lundberg2017unified,datta2016algorithmic}.

\subsection{Explaining system decisions}

As mentioned, our study focuses on AI systems that make, support, or recommend discrete decisions. Discrete decision making is closely related to classification, and the subtle distinction can often be overlooked safely. However, for explaining system decisions, it is important to be clear. First, there is a definitional difference: a classification model might classify someone as defaulting on credit or not; a corresponding decision-making system would use this model to decide whether to grant credit. Deciding not to grant credit is not the same (at all) as saying that the individual will default, which brings us to the technical difference.  

\textbf{Classification tasks} usually are modeled as scoring problems, where we want predictive models to score the observations such that those more likely to have the ``correct'' class will have higher scores. For example, scores may correspond to estimated probabilities of defaulting on credit, and individuals may be classified as defaulting or not based on their probability of defaulting. 

\textbf{Decision making} may also be modeled as a ``classification task'' by associating a class with each possible course of action (e.g., ``grant credit'' and ``do not grant credit''), which is related to (but usually not the same as) classifying individuals according to labels in the data. For example, estimated probabilities of class membership are often combined with application-specific information on costs and benefits to produce a next stage of more nuanced scores (e.g., the expected profits of granting credit). A system may then use these scores to make decisions using a chosen threshold appropriate for the problem at hand~\citep{provost2013data}. Thus, a credit scoring system may decide to extend credit to an individual with a relatively high probability of default if the interest rate is high, for example. 

Critically, this implies that the final output of the system (i.e., the decision) may not correspond to the labels in the training data. As an additional example, for a system deciding whether to target a customer with a promotion, scores could consist of expected profits. In this case, we could estimate a classification model to predict the probability that the customer will make a purchase and a regression model to estimate the size of the purchase (conditioned on the customer making a purchase); the expected profits would be the multiplication of these two predictions, and the ranking of the customers by expected profit could be different from the ranking based simply on the classification model score. The final output of the decision-making system would be whether the customer should be targeted with a promotion, which is not the same as predicting whether a customer will make a purchase (and because of selection bias and other complications, we often patently would not want to learn models based on training data of who was targeted).  

Explaining the decisions made by intelligent systems has received both practical and research attention 
from the IS community 
for decades~\citep{gregor1999explanations}.~\cite{martens2014explaining} provide an overview of the IS literature that frames, motivates, and explains the importance of explaining system decisions for system adoption, improvement, and use. Notably, prior work shows that the ability for intelligent systems to explain their decisions is necessary for their effective use: when users do not understand the workings of an intelligent system, they become skeptical and reluctant to use it, even if the system is known to improve decision-making performance~\citep{arnold2006differential,kayande2009incorporating}. 

More recently, for example, a field study in a department of radiology showed that the use of AI systems slowed, rather than sped up, the radiologists’ decision-making process because the AI systems often provided recommendations that conflicted with the doctors’ judgment~\citep{lebovitz2019doubting}. Lacking critical understanding of the opaque AI systems, the doctors often relied on their own diagnoses, which did not concur with the system's. This result highlights the need for methods to make the decisions of such AI systems more transparent.



\section{Counterfactual Explanations}


We now present counterfactual explanations for system decisions in detail, showing where we generalize from prior work. In general, counterfactual explanations describe a situation in the form: ``if X had not occurred, Y would not have occurred.'' Thus, counterfactual explanations can be used to explain system behavior in a narrow causal sense.\footnote{We do not intend to add to the philosophy or long debate on counterfactual theories of causation \citep{menzies2001counterfactual}.}  Specifically: What system input caused the system to make the decision that it did? Understanding what causes computer system decisions is easier than understanding the causality of natural phenomena. It is easier because we can directly observe the counterfactual ``if A had not occurred, B would not have occurred'' by changing the system inputs and observing the system output. This does not necessarily say anything about causal relationships outside the computer system, such as in the data-generating process. The ``outcome'' (B) is the focal decision made by the system, and the ``causes'' (A) are the data inputs that drive the system to make the decision. Therefore, in our context, a counterfactual explanation consists of a set of data inputs that, when changed, results in a different system decision. For instance, in credit scoring, one could explain a credit denial decision by saying, ``if the applicant had not had such a low income, the system would have granted credit.''  


The idea of taking a causal perspective to explain system decisions with counterfactuals was first proposed (to our knowledge) in \textit{MIS Quarterly}~\citep{martens2014explaining}, and subsequent work expanded on the ideas and methods for counterfactual approaches to explaining predictions and decisions~\citep{provost2014youtube,explanationssite,verma2020counterfactual}. \cite{martens2014explaining} define explanations in terms of input data that would change the decision if it were not present. Unfortunately, that paper did not seem to make clear the general nature of the counterfactual explanations because the explanations were originally presented for document classification. While the framework  subsequently has been used in other business settings~\citep{provost2014youtube,moeyersoms2016explaining,chen2017enhancing,ramon2019counterfactual}, this initial use has led several researchers to state (mistakenly) in their work that the framework is specific to document classification and/or categorical features ~\citep[see, e.g.,][]{explanationssite,wachter2017counterfactual,biran2017explanation,tamagnini2017interpreting}.
We recast and generalize this framework to be more broadly applicable.

\subsection{Framing counterfactual explanations from an evidence-based perspective}

Counterfactual explanations are based on hypothetical realities that differ from the observed facts, but for these explanations to be useful, these realities must be plausible. This leads to three fundamental challenges. First, we must define what plausible means, which will vary across contexts. Second, searching for all potential explanations may be intractable. Third, there may be multiple explanations for each decision, so we may need to define criteria to choose (or rank) explanations. 

Like importance-weight methods that assess feature importance by simulating lack of knowledge about features, we argue that some of these challenges may be partially addressed by framing counterfactual explanations in terms of \textbf{absent evidence}: explanations may be framed in terms of features that change the system decision when the evidence they provide is no longer present. For illustration, suppose a credit card transaction was flagged for action by a data-driven AI system after it was registered as occurring outside the country where the cardholder lives, and suppose the system would not have flagged the transaction absent this location.\footnote{We should keep in mind the decision-rather-than-classification perspective. The decision is to flag the transaction for one or more actions, such as sending a message to the account holder to verify. Flagging may be based on a threshold on the estimated likelihood of fraud but may also consider the existence of evidence from other transactions and the potential loss if the transaction is fraudulent.} In this case, it is intuitive to consider the location of the transaction as an explanation for the system decision. There could be other explanations. Perhaps the transaction also involved a consumption category outside the cardholder's profile (e.g., a purchase at a casino), and excluding this information would also change the decision to ``do not flag.'' Both are counterfactual explanations---they comprise evidence without which the system would have made a different decision. 

This perspective offers several advantages when addressing the challenges mentioned above. First, absent evidence may be used to define a reasonable set of plausible changes. For instance, in the example above, ``removing evidence'' from a model-based decision-making procedure may imply replacing the location of the transaction with the country where the cardholder lives or replacing the consumption category with the cardholder's most common consumption category. Second, using absent evidence narrows  the set of potential explanations substantially because the point is not to consider all the different ways in which the features could be changed, but rather what would be the effect of not having some particular evidence. 
Third, we may rank explanations according to the relevance of the features in them (e.g., location may be easier to communicate than consumption category) . We discuss these advantages below, after formally defining counterfactual explanations using the evidence-based perspective.

Another subtle implication of this perspective is that its explanations are generally applied to ``non-default'' decisions because data-driven systems usually make default decisions in the absence of evidence suggesting that a different decision should be made. In our example, a transaction would be considered legitimate unless there is enough evidence suggesting fraud. As a result, explaining default decisions often corresponds to saying, ``because there was not enough evidence to make a non-default decision.''\footnote{This is not always the case. For example, if a credit card transaction was made in a foreign country, but the cardholder recently reported a trip abroad, the trip report could be a reasonable explanation for the transaction being classified as legitimate. So, the evidence in favor of a non-default decision may be canceled by evidence supporting a default decision.} Thus, we focus on explaining non-default decisions.

\subsection{Defining counterfactual explanations}\label{sec:definition}

Following \cite{martens2014explaining}, we define a counterfactual explanation for a system decision as a set of features that is \textbf{causal} and \textbf{irreducible}. Being causal means that removing the feature set or setting each feature in the set to some predetermined counterfactual value (e.g., the mean)\footnote{We will focus on setting features to counterfactual values for several reasons. First, simply removing features can be problematic for many AI systems, because they require values for input features, and thus require feature-value imputation if features are made “not present.” Second, setting specific counterfactual values can make more sense than removing features depending on the context of the application. Finally, the other methods in our comparisons use feature-value imputation to produce explanations, so doing the same allows an apples-to-apples comparison.  Nonetheless, our framework would also apply to simply removing features, under the condition that the AI system being explained can deal with feature removal.} causes the system decision to change.\footnote{As mentioned, it is critical to differentiate what is causing the data-driven system to make its decisions from causal influences in the actual data-generating processes in the ``real'' world. Counterfactual explanations for AI system decisions relate to the former and do not necessarily tell us anything about the latter.} Irreducible means that no proper subset of the explanation is causal. The importance of an explanation being causal is straightforward: the decision would have been different if not for the specific values of the features in the set (the ``evidence''). The irreducibility condition serves to avoid including superfluous features, which relates to the fact that some of the features in a causal set may not be necessary for the decision to change.

Formally, we define counterfactual explanations as follows. Consider an instance $I$ for which the decision-making system $C:I\rightarrow\{1,...,k\}$ gives the decision $c$. Instance $I$ consists of a set of features (or attributes) taking on specific values, such as {\tt income=\$50,000} or {\tt country=FRANCE}, and feature evidence is ``removed'' by setting the corresponding feature to some predetermined counterfactual value that makes sense in the particular application (e.g., the mean or the mode for the attribute in the population of interest). Then, given a set of features $E$ (e.g., \{{\tt income}, {\tt country}\}), $I'(E)$ represents instance $I$ after setting the features in $E$ to their respective counterfactual values, and $E$ is a counterfactual explanation for $C(I)=c$ if and only if:
\begin{gather}
    C(I'(E))\neq c~\text{(the explanation is causal)} \\
    \forall E' \subset E: C(I'(E'))= c~\text{(the explanation is irreducible)}
\end{gather}

This definition builds on the explanations proposed by Martens and Provost (2014), who developed and applied counterfactual explanations for document classifications.  In the context of document classification, an explanation was thus defined as an irreducible set of terms (e.g., words, phrases, n-grams) such that removing the set from a document changes its classification. Our definition generalizes their counterfactual explanations in two important ways. First, it makes explicit how the explanations may be used for broader system decisions. Second, their practical implementation of explanations removes evidence by setting features to zero, whereas we generalize to arbitrary counterfactual values.

\begin{figure}
    \footnotesize
    \centering
    \begin{tabular}{l}
    \toprule
     \textbf{Explanation 1:} Credit denied because  \{`loan\_amnt'\} is above average. \\
    \textbf{Explanation 2:} Credit denied because  \{`annual\_inc'\} is below average. \\
    \textbf{Explanation 3:} Credit denied because  \{`fico\_range\_high', `fico\_range\_low'\} are below average. \\
    \bottomrule 
    \end{tabular}
    \caption{Examples of counterfactual explanations for a system decision}
    \label{fig:counterfactual_intro}
\end{figure}

Going back to our credit scoring example, suppose a decision-making system using the model prediction explained in~Figure~\ref{fig:weights_intro} decides not to grant credit to that individual. Figure~\ref{fig:counterfactual_intro} shows some counterfactual explanations for the credit denial decision.

\subsection{Removing feature evidence}

A vital practical question raised by our definition of counterfactual explanations is what counterfactual values should be used to ``remove" feature evidence from instances? Most explanation methods, including methods that do not provide counterfactual explanations, such as importance-weight methods, typically simulate lack of knowledge about features by replacing their values with some default value, such as the mean. For example,~\cite{martens2014explaining} replace the value of features that represent the presence of a word in the document (binary indicator, count, term frequency–inverse document frequency (TFIDF) value, etc.) with a zero. This makes sense in the context of document classification because if we consider the presence of a word as evidence for a classification, removing that evidence---that word---would be represented by a zero for the corresponding feature.\footnote{They discuss the case where the absence of a word would be evidence as well; see the original paper.} However, different evidence removal strategies may be more appropriate in other applications, such as in the cardholder-level perspective discussed in our fraud example.  

The explanation framework we present is agnostic to which method is used to define the counterfactual values associated with each feature, taking the position that this decision is domain and problem dependent. For example,~\cite{saar2007handling} discuss various strategies for dealing with missing features when applying predictive models; any of them could be used in conjunction with this framework to define counterfactual values. In our comparison with feature-importance methods presented below, we use the same approach that those methods use (mean imputation) so that the analysis is a comparison of the two different types of methods and is not confounded by using different strategies for replacing feature values. Nevertheless, in one of the case studies, we illustrate model-based imputation as an alternative approach to demonstrate how it satisfies different needs when producing explanations. 

Importantly, within a particular domain and explanation context, the system developers and domain experts should choose the most appropriate method for producing counterfactual values, as a one-size-fits-all strategy is unlikely to work well in practice. The examples in this study are meant as a broad guideline of when certain methods can work better. For example, if a manager wants to understand why a certain credit application was rejected, then using mean imputation to explain that the application was rejected because the applicant's annual income is below average could be reasonable. However, this explanation may not suffice to the applicant, particularly if the explanation is meant to be used as a recommendation for how to get the credit approved. In such cases, model-based imputation could be used to develop counterfactual values that match those of similar applicants whose credit application was approved.

\subsection{A procedure for finding useful counterfactual explanations}\label{sec:heuristic}

Our definition of counterfactual explanations for system decisions allows any procedure for finding such explanations. For example, fast solvers for combinatorial problems may be used to find counterfactual explanations~\citep{schreiber2018optimal}. We adopt heuristic procedures in this paper. Algorithm~\ref{alg:1} shows how a generalization of the algorithm proposed by~\cite{martens2014explaining} may be used to find counterfactual explanations. Algorithm~\ref{alg:1} generalizes the original algorithm by using $I'(E)$ to represent instance $I$ after setting the features in $E$ to their respective counterfactual values. Feature values  were always set equal to zero in the original algorithm, which would be a specific instance of $I'(E)$ in our generalized framework. The second generalization is presented below: the introduction of a preference function.

This algorithm finds counterfactual explanations using a heuristic search that requires the decision to be based on a scoring function, such as a probability estimate from a predictive model. The search algorithm then uses this scoring function to first consider features that, when changed to their counterfactual values, reduce the score of the predicted class the most. This heuristic may be desirable when the goal is to find the smallest explanations, such as when explaining the decisions of models that use thousands of features. Another possible heuristic is to consider features according to their overall importance for the prediction, where the importance may be computed by a feature importance explanation technique~\citep{ramon2019counterfactual}. Both heuristics have been shown to scale well in high-dimensional settings~\citep{martens2014explaining,ramon2019counterfactual}.

\begin{algorithm}[ht!]
\setstretch{1}
\caption{Evidence-based explainer (EBE)}\label{alg:1} 
\SetAlgoLined
\LinesNumbered
\DontPrintSemicolon
\SetKwInOut{KwInput}{Input}
\SetKw{Break}{break}
\SetKwInOut{KwOutput}{Output}
\SetArgSty{textnormal}
\KwInput{$I=\{A_1=a_1,A_2=a_2,...,A_m=a_m\}$ \% Instance consisting of a set of features\newline $f_c:I\rightarrow \mathbf{R}$ \% Scoring function \newline $C:I\rightarrow \{1,2,...,k\}$ \% Decision-making system that uses scoring function $f_c$  \newline $max\_iteration = 30$ \% Maximum number of iterations}
\KwOutput{$explanation\_list$ \% List of explanations}
$c=C(I)$ \% The decision made by the system\;
$p=f_c(I)$ \% The score that led to decision $c$ \;
$explanation\_list=$ [ ] \;
$i=0$\;
$combinations= $ Initialize empty list of sets ordered by scores (from lowest to largest) \;
Insert an empty set with score $p$ to $combinations$ \;
\While{$i<max\_iteration$ AND $combinations$ is not empty}{
    $combination$ = pop set with the smallest score from $combinations$\;
    $counterfactual = I'(combination)$ \% Set features to their counterfactual values\;
    \eIf{$C(counterfactual) = c$}{
        \ForEach{feature $A_j$ in $counterfactual$ with a non-counterfactual value}{
            $E = combination \cup \{A_j\}$ \;
            \If{$E$ is not a superset of an explanation in $explanation\_list$}{
                $p = f_c(I'(E))$ \;
                Insert set $E$ with score $p$ to $combinations$ \;
            }
        }
    }{
        \% Entering here implies the combination is causal \;
        $explanation$ = $combination$ \;
        \% The following ensures the combination is irreducible \;
        \ForEach{set $E$ in the power set of $combination$}{
            \If{$C(I'(E)) \neq c$ AND $E$ is smaller than $explanation$}
            {$explanation$ = $E$}
        }
        Add $explanation$ to $explanation\_list$ \;
    }
    $i = i+1$\;
 }
\end{algorithm}

However, the shortest explanations are not necessarily the best explanations. For instance, users may want to use the explanations as guidelines for what to change in order to affect the system decision. As an example, suppose that a system decides to warn a man that he is at high risk of having a heart attack. An explanation that “the system would not have made the warning if the patient were not male” is of little use as a guide for action. Generally, some features will lead to better explanations than others depending on the application. For example, some features may be easy to change, while others may be practically impossible to change (e.g., gender)---so while an explanation including a very difficult-to-change feature would indeed explain the decision, it would not give practical guidance toward what could be done to affect the decision.

Therefore, we allow the incorporation of a preference function as part of the heuristic procedure to search first for the most relevant explanations. We pose the preference function as a cost function on the feature changes: the cost function associates costs to the adjustment of features so that sets of features that satisfy desirable characteristics are searched first. Importantly, the cost function is meant to be used as a mechanism to capture the relevance of explanations, so the cost of changing the features might not represent an actual cost (see the second case study). Returning to the heart attack example: if we assign an infinite cost to changing the gender feature, the heuristic would not select feature combinations that include it, regardless of its high impact on the output score. Instead, the heuristic would prefer explanations with many modest but “cheap” changes, such as changing several daily habits.

More specifically, we adjust the procedure proposed by~\cite{martens2014explaining} so that the heuristic searches first for the feature combinations for which the output score changes the most per unit of cost. Doing so only requires the definition of a cost function $c(I, E)$, which represents the ``cost'' of setting the features in $E$ to their respective counterfactual values, and sorting potential explanations in descending order according to $(f_c(I)-f_c(I'(E))/c(I, E)$ (i.e., the score change per unit of cost) instead of in ascending order according to $f_c(I'(E))$ (as in lines 14-15 of Algorithm~\ref{alg:1}). Similar approaches have been suggested for inverse classification~\citep{lash2017generalized}, any of which could be used to find counterfactual explanations.

\section{Limitations of Importance Weights}

This section uses three simple, synthetic and illustrative examples to highlight two reasons importance-weight explanations may not be well suited to explain data-driven decisions made by AI systems. Example 1 illustrates that features with a large impact on a prediction (and thus large importance weights) may not affect the decision made using that prediction. The next two examples show that importance weights are insufficient to communicate how features actually affect decisions (even when importance is determined with respect to system decisions rather than model predictions). More specifically, we show that importance weights can remain the same despite substantial changes to decision making (Examples 1, 2, and 3) and that features deemed unimportant by the weights can actually affect the decision (Example 3). Similar examples to the ones discussed in this section will come up again in the case studies in Section~\ref{sec:cases} when comparing importance weights with counterfactual explanations using real-world data.

Throughout this section, the examples assume that we want to explain the binary decision made for a three-feature instance $I$ and decision procedure $C_i$ as defined here:
\begin{gather}
    I=\{A_1=1,A_2=1,A_3=1\} \\
    C_i(I)=\begin{cases}
    1,& \text{if } \hat{Y}_i(I)\geq 1\\
    0,              & \text{otherwise}
\end{cases}
\end{gather}
where $\{A_1,A_2,A_3\}$ are binary attributes, and $C_i$ is the decision-making procedure (an AI system) that employs the scoring (or prediction) function $\hat{Y}_i$ to make decisions. The examples that follow will use different $\hat{Y}_i$. The examples assume that domain knowledge has guided us to set feature values equal to zero when considering features as part of a counterfactual explanation.

We compute importance weights using SHAP~\citep{lundberg2017unified}, a popular approach to explain the output of machine learning models. Before we focus on the disadvantages of importance weights for explaining system decisions, let us point out that SHAP has several advantages for explaining data-driven model predictions: (i) it produces numeric ``importance weights'' for each feature at an instance level, (ii) it is model agnostic, (iii) its importance weights tie instance-level explanations to  cooperative game theory, providing a solid theoretical foundation, (iv) and SHAP unites several feature importance weighting methods, including the relatively well known LIME (Ribeiro, Singh and Guestrin, 2016).

In the case of SHAP, importance weights consist of the (approximated) Shapley values of the features for a model prediction. Shapley values correspond to the impact each feature has on the prediction, averaged over all possible joining orders of the features. In this context, a joining order is a permutation according to which the impact of the features on a model's prediction is considered (e.g., first $A_2$, then $A_3$, and lastly $A_1$). The impact of a feature corresponds to the change in the model prediction when the feature's default (counterfactual) value is replaced with the value observed for that instance, and the Shapley value consists of the average impact across permutations. We illustrate the computation of Shapley values precisely in the examples below.

A major limitation of Shapley values is that computing them becomes intractable as the number of features grows. SHAP circumvents this limitation by sampling the space of joining orders, resulting in a sampling-based approximation of the Shapley values. There are only three features  in the examples that follow, so the approximations are not necessary here, but they will be needed for the case studies presented in Section~\ref{sec:cases}, where the number of features is much larger.

\subsection{Example 1: Distinguishing between predictions and decisions}

All importance weighting methods (that we are aware of) are designed to explain the output of scoring functions, not system decisions. This is problematic because a large impact on the scoring function does not necessarily translate to an impact on the decision. Example 1 illustrates this point by defining $\hat{Y}_1$ as follows:
\begin{equation}
    \hat{Y}_1(I)=A_1+A_2+10A_1A_3+10A_2A_3,
\end{equation}
so the prediction and the decision for instance $I$ are $\hat{Y}_1(I)=22$ and $C_1(I)=1$, respectively. Note that in practice, we often do not know the exact functional form of $\hat{Y}_i$ or $\hat{C}_i$, so we are not able to peek into the internals of the explained model/system and can only probe it by feeding it different inputs and examining the outputs.

Table~\ref{tab:example1y} shows how to compute the Shapley values of the features with respect to $\hat{Y}_1$. Each row represents one of the six possible joining orders of the features, and each column corresponds to the impact of one of the three features across those joining orders. The last row shows the average impact of the features across the joining orders, which corresponds to the Shapley values.

\renewcommand{\arraystretch}{1.5}

\begin{table}
    \centering
    \begin{tabular}{c|c|c|c}
        \toprule
         \textbf{Joining orders}&\textbf{Impact of $A_1$}&\textbf{Impact of $A_2$}&\textbf{Impact of $A_3$}  \\
         \midrule
         $A_1,A_2,A_3$& 1 & 1 & 20 \\ \hline
         $A_1,A_3,A_2$& 1 & 11 & 10 \\ \hline
         $A_2,A_1,A_3$& 1 & 1 & 20 \\ \hline
         $A_2,A_3,A_1$& 11 & 1 & 10 \\ \hline
         $A_3,A_1,A_2$& 11 & 11 & 0 \\ \hline
         $A_3,A_2,A_1$& 11 & 11 & 0 \\ \midrule
         \textbf{Shapley values} & 6 & 6 & 10 \\ 
         \bottomrule
         
    \end{tabular}
    \caption{Shapley values for $\hat{Y}_1$ and all the joining orders used in their computation.}
    \label{tab:example1y}
\end{table}

According to Table~\ref{tab:example1y}, SHAP gives $A_3$ a larger weight than $A_1$ or $A_2$ due to its large impact on $\hat{Y}_1$. However, if we take a closer look at $C_1$ and $\hat{Y}_1$ simultaneously, we can see that $A_3$ does not affect the decision-making procedure at all! More specifically, $A_3$ only affects $\hat{Y}_1$ if $A_1$ or $A_2$ are already present, but if those features are present, then increasing the score does not affect the decision because $\hat{Y}_1$ is already greater than or equal to one (implying that $C_1=1$ regardless of $A_3$). Therefore, the large ``importance'' of a feature for a model prediction may not imply any impact on a decision made with that prediction.

We might then conclude that the issue would be solved by using SHAP to compute feature importance weights for system decisions (rather than for model predictions). Table~\ref{tab:example1c} shows the Shapley values of the features with respect to the decision-making procedure $C_1$ instead of $\hat{Y}_i$.\footnote{SHAP can also be used to explain non-binary categorical decisions by transforming the output of the decision system into a ``scoring function'' that returns $1$ if the decision is the same after changing the features and returns $0$ otherwise. This transformation, originally introduced by~\cite{moeyersoms2016explaining} (also in the context of using Shapley values for instance-level explanations), would allow us to use SHAP to obtain importance weights even for decisions with multiple, unordered alternatives that cannot normally be represented as a single numeric score.} It illustrates that $A_3$ indeed does not affect the decision at all. However, the next examples show that, even when importance weights are computed with respect to the decision-making procedure rather than the model predictions, the weights do not effectively capture how features affect decisions .

\begin{table}
    \centering
    \begin{tabular}{c|c|c|c}
        \toprule
         \textbf{Joining orders}&\textbf{Impact of $A_1$}&\textbf{Impact of $A_2$}&\textbf{Impact of $A_3$}  \\
         \midrule
         $A_1,A_2,A_3$& 1 & 0 & 0 \\ \hline
         $A_1,A_3,A_2$& 1 & 0 & 0 \\ \hline
         $A_2,A_1,A_3$& 0 & 1 & 0 \\ \hline
         $A_2,A_3,A_1$& 0 & 1 & 0 \\ \hline
         $A_3,A_1,A_2$& 1 & 0 & 0 \\ \hline
         $A_3,A_2,A_1$& 0 & 1 & 0 \\ \midrule
         \textbf{Shapley values} & 0.5 & 0.5 & 0 \\
         \midrule
         \multicolumn{4}{l}{There is a single counterfactual explanation: $\{A_1, A_2\}$}\\
         \bottomrule
         
    \end{tabular}
    \caption{Shapley values for $C_1$ and all counterfactual explanations for this decision.}
    \label{tab:example1c}
\end{table}

\subsection{Example 2: Multiple interpretations for the same weights}\label{sec:example2}

In Example 1, the decision changes when we change $A_1$ and $A_2$ simultaneously, and changing any of the features individually does not change the decision. So, according to our definition in Section~\ref{sec:definition}, there is a single counterfactual explanation, $\{A_1,A_2\}$.  However, suppose we were to use the following scoring function to make decisions instead:
\begin{equation}
    \hat{Y}_2 = A_1A_2
\end{equation}

\begin{table}
    \centering
    \begin{tabular}{c|c|c|c}
        \toprule
         \textbf{Joining orders}&\textbf{Impact of $A_1$}&\textbf{Impact of $A_2$}&\textbf{Impact of $A_3$}  \\
         \midrule
         $A_1,A_2,A_3$& 0 & 1 & 0 \\ \hline
         $A_1,A_3,A_2$& 0 & 1 & 0 \\ \hline
         $A_2,A_1,A_3$& 1 & 0 & 0 \\ \hline
         $A_2,A_3,A_1$& 1 & 0 & 0 \\ \hline
         $A_3,A_1,A_2$& 0 & 1 & 0 \\ \hline
         $A_3,A_2,A_1$& 1 & 0 & 0 \\ \midrule
         \textbf{Shapley values} & 0.5 & 0.5 & 0 \\ \midrule
         \multicolumn{4}{l}{There are two counterfactual explanations: $\{A_1\}\text{ and } \{A_2\}$}\\
         \bottomrule
         
    \end{tabular}
    \caption{Shapley values for $C_2$ and counterfactual explanations for this decision.}
    \label{tab:example2}
\end{table}

Table~\ref{tab:example2} shows the Shapley values for $C_2$, which are the same as for $C_1$ (see Table~\ref{tab:example1c}) because features $A_1$ and $A_2$ are equally important in both cases. However, the decision-making procedure is different because the new scoring function implies that changing either feature would change the decision. Therefore, with the new scoring function, there would be two counterfactual explanations, $\{A_1\}$ and $\{A_2\}$, but the importance weights do not capture this. This implies that importance weights do not effectively communicate how changing the features may change the decision.

\subsection{Example 3: Positive impact of non-positive weights}\label{sec:example3}

Example 1 showed that even if a feature has a large, positive importance weight for a model's instance-level prediction, changing the feature may not affect the decision made for that instance.
Importance weights can also be misleading when used to explain system decisions because the opposite can also be true: a feature with an importance weight of zero may affect the decision. We illustrate this with a third example, for which we use the following scoring function:
\begin{equation}
\hat{Y}_3 = A_1+A_2-2A_1A_2-A_1A_3-A_2A_3+3A_1A_2A_3
\end{equation}

\begin{table}
    \centering
    \begin{tabular}{c|c|c|c}
        \toprule
         \textbf{Joining orders}&\textbf{Impact of $A_1$}&\textbf{Impact of $A_2$}&\textbf{Impact of $A_3$}  \\
         \midrule
         $A_1,A_2,A_3$& 1 & -1 & 1 \\ \hline
         $A_1,A_3,A_2$& 1 & 1 & -1 \\ \hline
         $A_2,A_1,A_3$& -1 & 1 & 1 \\ \hline
         $A_2,A_3,A_1$& 1 & 1 & -1 \\ \hline
         $A_3,A_1,A_2$& 0 & 1 & 0 \\ \hline
         $A_3,A_2,A_1$& 1 & 0 & 0 \\ \midrule
         \textbf{Shapley values} & 0.5 & 0.5 & 0 \\ \midrule
         \multicolumn{4}{l}{There are three counterfactual explanations: $\{A_1\}, \{A_2\}\text{, and } \{A_3\}$}\\
         \bottomrule
         
    \end{tabular}
    \caption{Shapley values for $C_3$ and counterfactual explanations for this decision.}
    \label{tab:example3}
\end{table}

Table~\ref{tab:example3} shows the Shapley values with respect to $C_3$. The values are the same as in the previous examples, but the decision-making process has changed once again. Notably, changing $A_3$ can change the decision from $C_3=1$ to $C_3=0$, as evidenced by the impact of $A_3$ in the first and third joining orders, but the importance weight of $A_3$ is $0$. The counterfactual explanation framework, on the other hand, reveals that there are three counterfactual explanations in this example: $\{A_1\}$, $\{A_2\}$, and $\{A_3\}$. Thus, a feature that we might mistakenly deem as irrelevant due to its non-positive weight is, in fact, just as important as the other features with positive weights for the purposes of explaining the decision $C_3 (I)=1$.

\subsection{Drawbacks of using averages}

While the previous examples were constructed to illustrate the limitations of importance weights, they reveal an important insight: it is difficult to capture the impact of features on decisions with a single number, especially when features interact with each other. This is particularly relevant when explaining black-box models (such as neural networks), which are well-known for learning complex interactions between features. Moreover, Section~\ref{sec:cases} shows how the hypothetical examples in this section also occur in real-world scenarios. 

The main reason importance weights are problematic for explaining system decisions is that they aggregate across potential explanations (i.e., feature sets) to provide a single explanation per decision. Thus, each decision is explained using a single vector of weights. Typically, the importance weighting methods summarize the impact of features in a single vector by averaging across multiple feature orderings.  The problem is that the average impact of a feature is not fine-grained enough to reveal dynamics between features. More importantly, it is difficult to interpret: why should the average across feature orderings be relevant to explain a decision? After all, it may not be representative of the potential impact of features (as in the case of $A_3$ in Example 3).

Counterfactual explanations circumvent the drawbacks of using averages because the explanations are defined at the counterfactual level, meaning that each explanation
represents a counterfactual world in which the decision would be different. This allows a single decision to have multiple explanations, allowing a richer interpretation of how the features may influence the decision. Table~\ref{tab:weight_counterfactual_comparison} summarizes the differences between the two approaches.

\begin{table}[ht!]
    \small
    \centering
    \begin{tabular}{c|p{5.3cm}|p{5.3cm}}
    \toprule
    &\multicolumn{1}{c|}{\textbf{Importance weights}}& 
    \multicolumn{1}{c}{\multirow{2}{*}{\textbf{\shortstack{Counterfactual explanations \\(evidence-based framework)}}}}\\
    &&\\
         \midrule
         \textbf{Unit of analysis} & Instance level. There is one explanation for each system decision. & Counterfactual level. There may be multiple or no explanations for each system decision.  \\
         \midrule
         \textbf{Output explained} & Model predictions (although the methods can be adapted to explain system decisions). Critically, a feature that affects model predictions may not affect system decisions. & System decisions.  \\
         \midrule
         \textbf{Design Intent} & Quantify feature importance. The explanations do not communicate how system decisions change as a result of changing the features. & Explain system decisions. The explanations are defined in terms of how system decisions change as a result of changing features.  \\
         \midrule
         \textbf{Approach} & Summarize the impact of each feature in a single number. However, features may affect system decisions in different ways depending on the values of other features. & Identify features that affect the system decision within the context of specific values for the other features.  \\
    \bottomrule
    \end{tabular}
    \caption{Summary of differences between importance weights and counterfactual explanations to explain system decisions}
    \label{tab:weight_counterfactual_comparison}
\end{table}

\section{Case Studies}\label{sec:cases}

We now present three case studies to illustrate these phenomena using real-world data.\footnote{The code is available but is blinded for review.} The first case study contrasts counterfactual explanations with explanations based on importance weights, showing fundamental differences. The second case study showcases the power of counterfactual explanations for high-dimensional data and shows how the heuristic procedure that generates counterfactual explanations may be adjusted to search and sort explanations according to their relevance to the decision maker. The third case study shows the application of counterfactual explanations to AI systems that are more complex than simply applying a threshold to the output of a single predictive model, specifically, to systems that integrate multiple models predicting different things. In all case studies, we use SHAP to compute importance weights with respect to the decision-making procedure rather than model predictions (as discussed above). 

\subsection{Study 1: Importance Weights vs. Counterfactual Explanations}

To showcase the advantages of counterfactual explanations over feature importance weights when explaining data-driven decisions, we explain system decisions to accept or deny credit based on real data from Lending Club, a peer lending platform. The data contain comprehensive information on all loans issued starting in 2007. The data set includes hundreds of features for each loan, including interest rate, loan amount, monthly installment, loan status (e.g., fully paid, charged off), and other attributes related to the borrower, such as type of home ownership and annual income. To simplify the setting, we use the data sample used by~\cite{cohen2018data} and focus on loans with a 13\% annual interest rate and a duration of three years (the most common loans), resulting in 71,938 loans. The loan decision making is simulated but is in line with consumer credit decision making as described in the literature~\citep[see][]{baesens2003benchmarking}.\footnote{The Lending Club data contain a substantial number of loans for which traditional models estimate moderately high likelihoods of default, despite these all being issued loans. This may be due to Lending Club’s particular business model, where external parties choose to fund (invest in) the loans.}

We use 70\% of this data set to train a logistic regression model that predicts the probability of borrowers defaulting using the following features: loan amount (loan\_amnt), monthly installment  (installment), annual income (annual\_inc), debt-to-income ratio (dti), revolving balance (revol\_bal), incidences of delinquency (delinq\_2yrs), number of open credit lines (open\_acc), number of derogatory public records (pub\_rec), upper boundary range of FICO score (fico\_range\_high), lower boundary range of FICO score (fico\_range\_low), revolving line utilization rate (revol\_util), and months of credit history (cr\_hist). The model is used by a simulated system that denies credit to loan applicants with a probability of default above 23\%. We use the system to decide which of the held-out 30\% of loans should be approved.

\begin{figure}
    \centering
    \includegraphics[width=1\textwidth]{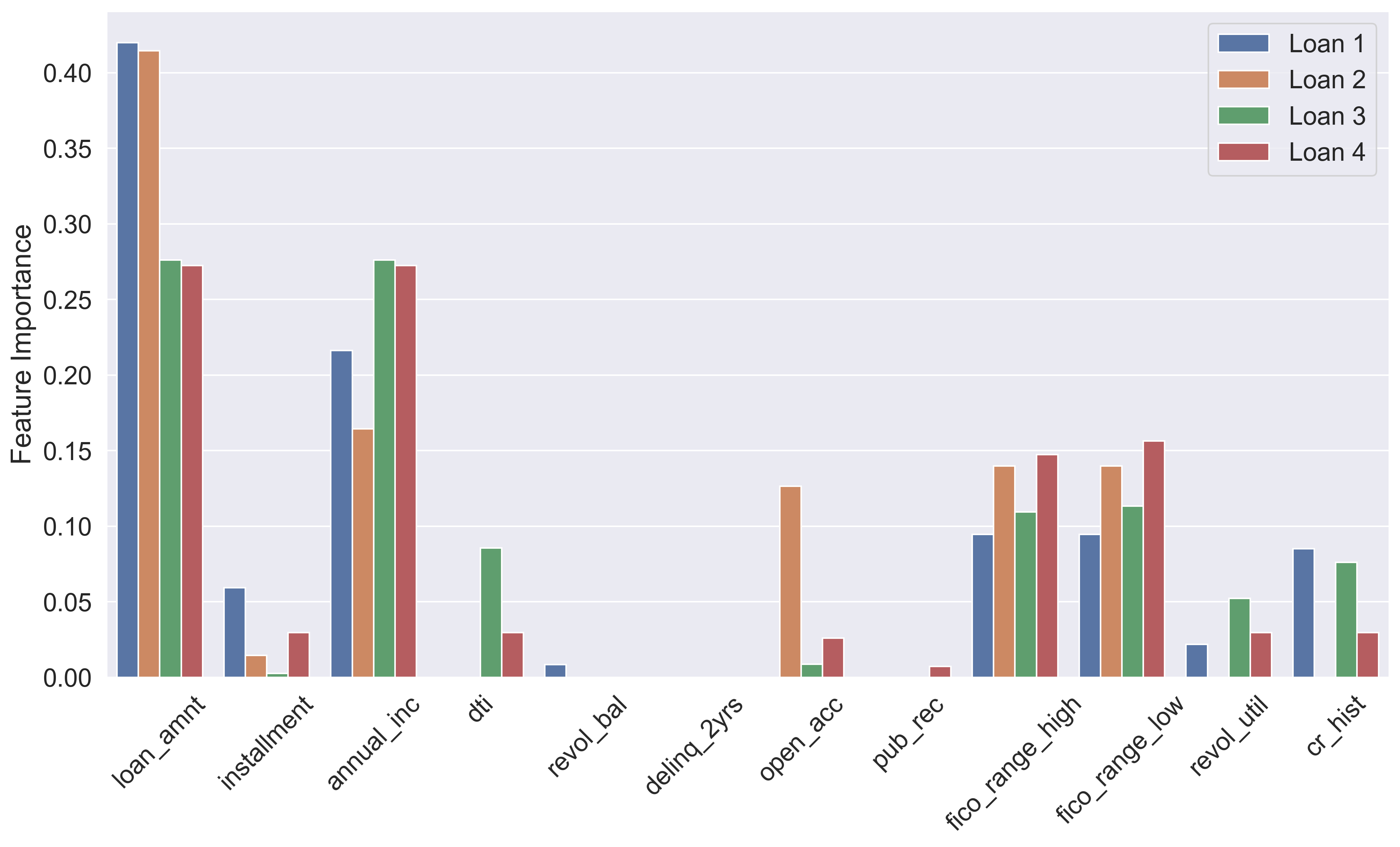}
    \caption{Feature importance weights according to SHAP}
    \label{fig:weights}
\end{figure}

Comparing counterfactual explanations to explanations based on feature importance weights shows that counterfactual explanations have several advantages. First, importance weights do not communicate which features would need to change in order for the decision to change, so their role as explanations for decisions is incomplete. Figure~\ref{fig:weights} shows the feature importance weights assigned by SHAP to four loans (different colors) denied by the system. For instance, according to SHAP, loan\_amnt was the most important feature for the credit denial of all four loans. However, this information does not fully explain any of the decisions. The credit applicant of Loan 1, for example, cannot use the explanation to understand what would need to be different to obtain credit.  Was it the amount of the loan? The annual income? Both?

Table~\ref{tab:loan1}, in contrast, shows all counterfactual explanations for the credit denial decision of Loan 1. Each column represents an explanation, and the arrows in each cell show which features are present in each explanation (recall that a counterfactual explanation is a set of features). The last column shows the difference between the original value of each feature and the value that was imputed to simulate evidence removal (the mean in this case), illustrating how our generalized counterfactual explanations may be applied to numeric features. 

\begin{table}[ht!]
    \centering
    \begin{tabular}{c|c|c|c|c|c|c|c}
    \toprule
    \multirow{2}{*}{\textbf{Features}}
         & \multicolumn{6}{c|}{\textbf{Explanations}} & \multirow{2}{*}{\textbf{\shortstack{Distance\\ from mean}}}\\
    \hhline{~------~}
    &\textbf{1}&\textbf{2}&\textbf{3}&\textbf{4}&\textbf{5}&\textbf{6}& \\
    \midrule
    loan\_amnt & $\uparrow$ &&&&&&+\$16,122 \\
    \hline
    installment &&&&&$\uparrow$&&+\$540 \\
    \hline
    annual\_inc && $\downarrow$& $\downarrow$& $\downarrow$& $\downarrow$& $\downarrow$&-\$9,065 \\
    \hline
    dti &&&&&&&n/a \\
    \hline
    revol\_bal &&&&&& $\downarrow$& -\$4,825 \\
    \hline
    delinq\_2yrs &&&&&&&n/a \\
    \hline
    open\_acc &&&&&&&n/a \\
    \hline
    pub\_rec &&&&&&&n/a \\
    \hline
    fico\_range\_high &&&$\downarrow$&&&&-16\\
    \hline
    fico\_range\_low &&$\downarrow$&&&&&-16 \\
    \hline
    revol\_util &&&&&&$\uparrow$&+12\% \\
    \hline
    cr\_hist &&&&$\downarrow$&&&-92 months \\
    \midrule
    \multicolumn{8}{l}{$\uparrow$ means feature is \textbf{too large} to grant credit.}\\
    \multicolumn{8}{l}{$\downarrow$ means feature is \textbf{too small} to grant credit.} \\ 
    \bottomrule
    \end{tabular}
    \caption{Counterfactual explanations for Loan 1}
    \label{tab:loan1}
\end{table}

For example, as shown in column 1, one possible explanation for the credit denial of Loan 1 is that the loan amount is too large (it is \$16,122 larger than the average), given the other aspects of the application. One could explain the decision in several other ways. The explanation in column 4 suggests that the \$28,000 loan would be approved if the applicant's annual income and credit history had not been below average. These explanations make it immediately apparent how the features influenced the decision. This highlights two additional advantages of counterfactual explanations: they give a deeper insight into why the credit was denied and provide alternatives that could change the decision.

\begin{table}[ht!]
    \centering
    \resizebox{\textwidth}{!}{
    \begin{tabular}{c|c|c|c|c|c|c|c|c|c|c|c|c|c|c|c|c}
    \toprule
    \multirow{2}{*}{\textbf{Features}}
         & \multicolumn{15}{c|}{\textbf{Explanations}} & \multirow{2}{*}{\textbf{\shortstack{Distance \\ from mean}}}\\
    \hhline{~---------------~}
    &\textbf{1}&\textbf{2}&\textbf{3}&\textbf{4}&\textbf{5}&\textbf{6}&\textbf{7}&\textbf{8}&\textbf{9}&\textbf{10}&\textbf{11}&\textbf{12}&\textbf{13}&\textbf{14}&\textbf{15}& \\
    \midrule
    loan\_amnt & $\uparrow$ &&&&&&&&&&&&&&&+\$16,122 \\
    \hline
    installment &&&&&&$\uparrow$&&&&&&$\uparrow$&&&$\uparrow$&+\$540 \\
    \hline
    annual\_inc && $\downarrow$&&&&&&&&&&&&&&-\$9,065 \\
    \hline
    dti &&&&$\uparrow$&&&&&&$\uparrow$&&&&&$\uparrow$&+5 \\
    \hline
    revol\_bal &&&&&&&&&&&&&&&& n/a \\
    \hline
    delinq\_2yrs &&&&&&&&&&&&&&&& n/a \\
    \hline
    open\_acc &&&&&&&&$\uparrow$&&&&&&$\uparrow$&&+1 \\
    \hline
    pub\_rec &&&&&&&&&$\uparrow$&&&&&&&+1 \\
    \hline
    fico\_range\_high &&&$\downarrow$&&&&&&&$\downarrow$&$\downarrow$&$\downarrow$&$\downarrow$&$\downarrow$&&-16\\
    \hline
    fico\_range\_low &&&$\downarrow$&$\downarrow$&$\downarrow$&$\downarrow$&$\downarrow$&$\downarrow$&$\downarrow$&&&&&&&-16 \\
    \hline
    revol\_util &&&&&&&$\uparrow$&&&&&&$\uparrow$&&$\uparrow$&+12\% \\
    \hline
    cr\_hist &&&&&$\downarrow$&&&&&&$\downarrow$&&&&$\downarrow$&-92 months \\
    \midrule
    \multicolumn{17}{l}{$\uparrow$ means feature is \textbf{too large} to grant credit.}\\
    \multicolumn{17}{l}{$\downarrow$ means feature is \textbf{too small} to grant credit.} \\
    \bottomrule
    \end{tabular}}
    \caption{Counterfactual explanations for Loan 4}
    \label{tab:loan4}
\end{table}

Table~\ref{tab:loan4} shows the counterfactual explanations for Loan 4 to emphasize this last point. Figure~\ref{fig:weights} shows that the most important features for Loan 1 and Loan 4 are the same. From this figure alone, one may conclude that these two credit denial decisions should have similar counterfactual explanations. Yet, comparing Table~\ref{tab:loan1} and Table~\ref{tab:loan4} reveals that this in fact is not the case. Loan 4 has many more explanations, and even though the explanations in both loans have similar features, the only explanation that the loans have in common is the first (i.e., the loan amount is too large); there is no other match. 

Nonetheless, one can see that not all features shown in Figure~\ref{fig:weights} and Tables~\ref{tab:loan1}-\ref{tab:loan4} would be relevant for loan applicants looking for recommendations for how they might get their credit approved. SHAP may be adjusted further to compute weights only for a subset of features. Because SHAP also deals  with evidence removal by imputing default values, we can easily extend SHAP to only consider certain (relevant) features by setting the default values of the irrelevant features equal to the current values of the instance. Then, SHAP will compute importance weights only for the features with a value different from the default. We do this for Loan 4 and define loan amount and annual income as the only relevant features. This would make sense in our context if customers can only ask for less money or show additional sources of income.

Under these conditions, SHAP computes an importance weight of 0.5 for both the loan amount and the annual income, and there are two counterfactual explanations: the loan amount is too high, or the annual income is too low (columns 1 and 2 in Table~\ref{tab:loan4}). However, consider a different scenario. Suppose the bank were stricter with the loans it approves and used a decision threshold 2.5 percentage points lower. Now, the Loan 4 applicant would need both to reduce the loan amount and to increase their annual income before the loan would be approved.
This situation is directly analogous to Example 2 in Section~\ref{sec:example2}.
With this different decision system, there is a single counterfactual explanation (instead of two) consisting of both features, so the counterfactual framework captures the change in the decision-making procedure. However, SHAP would still show an importance weight of 0.5 for each feature. Thus, the counterfactual explanations and the SHAP explanations exhibit different behavior. SHAP explanations suggest that the two decisions are essentially the same. The counterfactual explanations suggest that they are quite different, which we argue is preferable in most settings. 

Another crucial aspect of counterfactual explanations is the method used to choose counterfactual values.\footnote{This aspect is also crucial for feature importance methods that use imputation to simulate lack of knowledge about features, such as SHAP. We are not aware of a similar discussion in the literature about those methods.} Such methods should be carefully chosen according to the domain and the problem. For example, mean imputation may be adequate to explain to a Lending Club investor why she should not invest in a particular loan, but other imputation methods may be more appropriate for explaining the same decision to the credit applicant. For instance, if the applicant is a 20-year old requesting a loan to pay for tuition, then having a short credit history may not be considered unusual. Thus, a more appropriate imputation method may consist of replacing the credit history with a value that is typical of individuals requesting student loans. 

We illustrate next how counterfactual explanations may be generated using model-based imputation. For each of the 12 available features, we fit a linear regression model using as training data the applicants who would be granted credit by the system and using the other features as predictors. We then use each of these models to impute feature values. For instance, in our previous example, the credit history model may be used to impute the expected credit history of the 20-year old requesting the student loan.

\begin{table}[ht!]
    \footnotesize
    \centering
    \begin{tabular}{c|c|c|c|c||c|c|c|c}
    \toprule
    \multirow{2}{*}{\textbf{Features}}
         & \multicolumn{4}{c||}{\textbf{Mean Imputation}} & \multicolumn{4}{c}{\textbf{Model-based Imputation}}\\
    \hhline{~--------}
    &\textbf{Loan 1}&\textbf{Loan 2}&\textbf{Loan 3}&\textbf{Loan 4}&\textbf{Loan 1}&\textbf{Loan 2}&\textbf{Loan 3}&\textbf{Loan 4}\\
    \midrule
loan\_amnt&16,122&15,722&6,672&12,372&123&79&-13&1\\
\hline
installment&539&528&226&417&-4&-3&1&0\\
\hline
annual\_inc&-9,065&-9,065&-27,065&-15,065&-32,568&-85,302&-29,903&-32,926\\
\hline
dti&-1&-1&10&4&2&-14&6&5\\
\hline
revol\_bal&-4,825&10,730&506&589&-9,750&-12,829&-10,982&-6,081\\
\hline
delinq\_2yrs&0&0&0&0&0&-1&0&0\\
\hline
open\_acc&-3&29&7&1&-4&26&6&-1\\
\hline
pub\_rec&0&0&0&1&0&0&0&1\\
\hline
fico\_range\_high&-16&-21&-26&-21&0&0&0&0\\
\hline
fico\_range\_low&-16&-21&-26&-21&0&0&0&0\\
\hline
revol\_util&11&-12&32&12&-3&6&22&1\\
\hline
cr\_hist&-92&-22&-104&-39&-91&-68&-113&-58\\
\bottomrule
    \end{tabular}
    \caption{Differences between observed values and default values for our four loans, using mean imputation and model-based imputation}
    \label{tab:default_values}
\end{table}

Table~\ref{tab:default_values} shows how the observed values for the four loans shown in Figure~\ref{fig:weights} differ from the corresponding default values when using mean imputation and model-based imputation (the entries in the table are these differences). The table shows several interesting results. First, although all the loan amounts are above average (according to mean imputation), these amounts are relatively common among other applicants with similar characteristics, as evidenced by the small differences with respect to the default value under model-based imputation. Therefore, even though the importance weights in Figure~\ref{fig:weights} hint at the loan amount as the primary reason for credit denial, this feature may not be considered relevant evidence in the context of these applicants. Moreover, the gap in annual income under model-based imputation is substantially larger, which reveals that the annual income of these applicants may be more anomalous than what is suggested by mean imputation. Additionally, features that consist of evidence against creditworthiness under mean imputation may actually be evidence in favor under model-based imputation (and vice versa). For example, in the case of Loan 4, the applicant has one more credit line than the average applicant, and as a result, this feature is part of two counterfactual explanations in Table~\ref{tab:loan4} (see open\_acc). However, according to model-based imputation, this user has one credit line less than other applicants with similar characteristics, which may be considered evidence in favor of creditworthiness depending on the context. Therefore, the method used to produce counterfactual values provides another way in which counterfactual explanations may be tailored to the context.

As an illustration, Table~\ref{tab:counterfactual_model_based} shows all the counterfactual explanations for the four loans shown in Figure~\ref{fig:weights} when using model-based imputation (rather than mean imputation). These explanations frame evidence against creditworthiness relative to  similar approved applications. The table shows that the counterfactual explanations have changed (compared to the ones in Tables~\ref{tab:loan1} and~\ref{tab:loan4}) to reflect that loan amount is no longer considered evidence against creditworthiness (because similar applicants have applied for similar amounts), whereas annual income is now considered the primary reason for credit denial.

\begin{table}[ht!]
    \centering
    \begin{tabular}{c|l}
    \toprule
     \textbf{Loan 1} & \textbf{Explanation 1.} Credit denied because: \\
     &- Annual income is \$32,568 less than expected. \\
    \hline 
    \textbf{Loan 2} & \textbf{Explanation 1.} Credit denied because: \\
     &- Annual income is \$85,302 less than expected. \\
    \hline 
    \textbf{Loan 3} & \textbf{Explanation 1.} Credit denied because: \\
     &- Annual income is \$29,903 less than expected. \\
                    & \textbf{Explanation 2.} Credit denied because: \\
     &- Debt-to-income ratio is 6 units more than expected. \\
     &- Revolving line utilization rate is 22\% more than expected. \\
     &- Credit history is 113 months less than expected. \\
                    & \textbf{Explanation 3.} Credit denied because: \\
     &- Debt-to-income ratio is 6 units more than expected. \\
     &- Open credit lines are 6 more than expected. \\
     &- Credit history is 113 months less than expected. \\
    \hline
     \textbf{Loan 4} & \textbf{Explanation 1.} Credit denied because: \\
     &- Annual income is \$32,926 less than expected. \\
    \bottomrule 
    \end{tabular}
    \caption{Counterfactual explanations using model-based imputation; ``expected'' refers to the value that model-based imputation predicts for this example.}
    \label{tab:counterfactual_model_based}
\end{table}

\subsection{Study 2: High-dimensional and Context-specific Explanations}

Our second case study uses Facebook data to showcase the advantages of counterfactual explanations when explaining data-driven decisions in high-dimensional settings. The data, collected through a Facebook application called myPersonality,\footnote{Thanks to the authors of the prior study,~\cite{kosinski2013private}, for sharing the data.} has also been used to compare the performance of various counterfactual explanation methods~\citep{ramon2019counterfactual}. We use a sample that contains information on 587,745 individuals from the United States, including their Facebook Likes and a subset of their Facebook profiles. In general, Facebook users do not reveal all their personal characteristics, but their Facebook Likes are available to the platform. To simulate a decision-making system for this case study, we assume a (fictitious) firm wants to launch a marketing campaign to promote a new product to users who are older than 50. Given that not all Facebook users share their age, the firm could use a predictive model to predict who is over 50 (using Facebook Likes) and use the predictions to decide whom to target with the campaign. 

A user’s Facebook Likes are the set of Facebook pages that the user chose to ``Like'' on the platform (we capitalize ``Like'', as have prior authors, to distinguish the act on Facebook). So, we represent each Facebook page as a binary feature that takes a value of 1 if the user Liked the page and 0 otherwise. We kept only the pages Liked by at least 1,000 users, leaving us with 10,822 binary features. The target variable for modeling is also binary and takes a value of 1 if the user is more than 50 years old and 0 otherwise. We use 70\% of the data to train a logistic regression model. In our fictitious setting, the model is used by a decision system that targets the top 1\% of users with the highest probability of being older than 50, which (in our sample) implies sending promotional content to the users with a probability greater than 41.1\%. We use the system to decide which of the held out 30\% of users to target.

Importantly, while the system could generate value for the firm, we need to consider a user's sense of privacy and how they might feel about being targeted with the promotional campaign. For example, some users may feel threatened by highly personalized offers (``How do they know this about me?'') and may want to know why they were targeted (see~\cite{chen2017enhancing} for a discussion). Explanations can also lead to higher user engagement resulting from
more confidence and transparency in product recommendations~\citep{friedrich2011taxonomy}. In such settings, users are unlikely to be interested in the intricacies of the model but rather in the data about their behavior used to target them. If that is the case, framing explanations in terms of comprehensible input features (e.g., Likes) is critical. 

One approach is to use importance weights to rank Facebook pages according to their feature importance (as computed by a technique such as SHAP) and then show the user the topmost predictive pages they Liked. However, given the large number of features (Facebook pages), computing weights in a deterministic fashion is intractable. SHAP circumvents this issue by sampling the space of feature combinations, resulting in sampling-based approximations of the influence of each feature on the prediction. However, the downside is that the estimated values may be far from the real values, which may yield inconsistent results. For example, if we were to use the topmost important features to explain a decision, we should consider whether different runs of a non-deterministic method repeatedly rank the same pages as the most important. Unfortunately, the set of the topmost important features becomes increasingly inconsistent (across different runs of SHAP) as the number of features increases. 

For instance, in our holdout data set, there is a 34-year-old user who would be targeted with an ad for older persons (the model predicts a 42\% probability that this user is at least 50 years old). So, as an example, suppose this user wants to know why they are targeted. Let's say that we have determined that showing the top-three most important features makes sense for this application. Table~\ref{tab:pages_shap} shows the top-three most predictive pages according to their SHAP values (importance weights) for the system decision. The table shows the result of running SHAP five times to compute the importance weights, each time sampling $4100$ observations of the space of feature combinations.\footnote{We use the SHAP implementation provided here: \url{https://github.com/slundberg/shap/}. 
At the moment of writing, the default sample size is $2048 + 2m$, where $m$ is the number of features with a non-default value. Our choice of $4100$ is larger than the SHAP implementation's default sample size for all of the experiments we run.} Because SHAP uses sampling-based approximations, we can see that SHAP values vary every time we compute them, resulting in different topmost predictive pages. Importantly, while some pages appear recurrently, only Paul McCartney appears in all five approximations.

\begin{table}
    \centering
    \resizebox{\textwidth}{!}{
    \begin{tabular}{c|c|c|c|c}
    \toprule
         \textbf{Approximation 1}&\textbf{Approximation 2}&\textbf{Approximation 3}&\textbf{Approximation 4}&\textbf{Approximation 5}  \\
         \midrule
         Elvis Presley& Paul McCartney& Paul McCartney& Paul McCartney& Elvis Presley\\
         (0.1446)& (0.1471) & (0.1823) & (0.1541) & (0.1582) \\
         \hline
         Bruce Springsteen& William Shakespeare & Neil Young & Elvis Presley & Paul McCartney\\
         (0.1302)& (0.1321) & (0.1676) & (0.1425) & (0.1489) \\
         \hline
         Paul McCartney& Brain Pickings & The Hobbit & Leonard Cohen & Bruce Springsteen\\
         (0.1268)& (0.1319) & (0.1417) & (0.1359) & (0.1303) \\
         \midrule
         \multicolumn{5}{l}{Importance weights (SHAP values) shown in parentheses.}\\
         \bottomrule
    \end{tabular}}
    \caption{Likes identified by SHAP as the three most important and their SHAP values for five different runs for a single decision to target our example user with the over-50 ad.}
    \label{tab:pages_shap}
\end{table}

As we will show in more detail below, this inconsistency is the consequence of using SHAP to estimate importance weights for too many features. This specific user Liked 64 pages, which is not an unusually large number of Likes---more than a third of the targeted users have at least that many. There are (at most) 64 non-zero SHAP values to estimate, making the task significantly simpler than if we had to estimate importance weights for all 10,822 features. However, even with a sample size already larger than the implementation's default, SHAP proves unreliable to find the most important pages (let alone to estimate the importance weights for each page). 
We increased the sample size for SHAP in order to observe when the estimates became stable for this particular task. For this specific user, it took eight times more samples from the feature space for the same three most-important pages to match consistently across all approximations, increasing computation time substantially (from 3 to 21 seconds per approximation on a standard laptop). This time would increase dramatically for data settings with hundreds of non-zero features, which are not uncommon~\citep[e.g., see][]{chen2017enhancing,perlich2014machine}.

In contrast, counterfactual explanations were found in a tenth of a second (on the same laptop), five of which we show in Figure~\ref{fig:pages_counterfactual}. Each explanation consists of a subset of Facebook pages that would change the targeting decision if removed from the set of pages Liked by the user. In other words, each of the sets shown in Figure~\ref{fig:pages_counterfactual} is an explanation in its own right, representing a minimum amount of evidence that (if removed) changes the decision. Importantly, these explanations are short, consistent (because they are generated in a deterministic fashion), and directly tied to the decision-making procedure.

\begin{figure}
    \small
    \centering
    \begin{tabular}{l}
    \toprule
     \textbf{Explanation 1:} The user would not be targeted if \{Paul McCarney\} had not been Liked. \\
    \textbf{Explanation 2:}  The user would not be targeted if \{Elvis Presley\} had not been Liked. \\
    \textbf{Explanation 3:} The user would not be targeted if \{Neil Young\} had not been Liked. \\
    \textbf{Explanation 4:}  The user would not be targeted if \{Leonard Cohen\} had not been Liked. \\
    \textbf{Explanation 5:} The user would not be targeted if \{Brain Pickings\} had not been Liked. \\
    \bottomrule 
    \end{tabular}
    \caption{Counterfactual explanations for a single decision to target our example user with the over-50 ad.}
    \label{fig:pages_counterfactual}
\end{figure}

As an additional systematic demonstration of the negative impact that an increasing number of features may have on the consistency of sampling-based feature-importance approximations, we show how the more pages a user has Liked, the more inconsistent the set of the top three most important pages becomes. The process is as follows. First, we picked a random sample of 500 users in the holdout data who would be targeted. Then, we applied SHAP five times to approximate the importance weights of the features used for each of the 500 targeting decisions (sampling 4,100 observations each time). Finally, for each targeting decision, we counted the number of pages that appeared consistently in the top three most important pages across all five approximations. We call this the number of matches. Thus, if the approximations were consistent, we would expect the same three pages to appear in the top three pages of all approximations, and there would be three matches. In contrast, if the approximations were completely inconsistent, no pages would appear in the top three pages of all five approximations, and there would be no matches. 

\begin{figure}
\centering
\begin{subfigure}{.49\textwidth}
    \centering
    \begin{tikzpicture}
\begin{axis}[
ymax=3,
ymin=0,
ymajorgrids=true,
xtick={1, 2, 3, 4, 5},
xticklabel style={align=center},
xticklabels={$<18$ \\Likes, 18--33 \\ Likes, 34--58 \\ Likes, 59--98  \\ Likes, $>98$ \\ Likes},
ytick distance={.5},
title = Average Matches (SHAP)
]
\addplot[color=blue]
    coordinates {(1, 2.71) (2, 2.25) (3, 2.14) (4, 1.71) (5, 1.15)};
\end{axis}
\end{tikzpicture}
    \caption{Average matches by quantile}
    \label{fig:matches}
\end{subfigure}
\begin{subfigure}{.49\textwidth}
    \centering
    \begin{tikzpicture}
\begin{axis}[
ymax=6,
ymin=0,
ymajorgrids=true,
xtick={1, 2, 3, 4, 5},
xticklabel style={align=center},
xticklabels={$<18$ \\Likes, 18--33 \\ Likes, 34--58 \\ Likes, 59--98 \\ Likes, $>98$ \\ Likes},
ytick distance={1},
title = Average Explanation Size
]
\addplot[color=blue]
    coordinates {(1, 1.17) (2, 2.0) (3, 2.30) (4, 3.60) (5, 4.90)};
\end{axis}
\end{tikzpicture}
    \caption{Average size by quantile}
    \label{fig:sizes}
\end{subfigure}
\caption{Variations in explanations by number of Likes}
\label{fig:exper}
\end{figure}
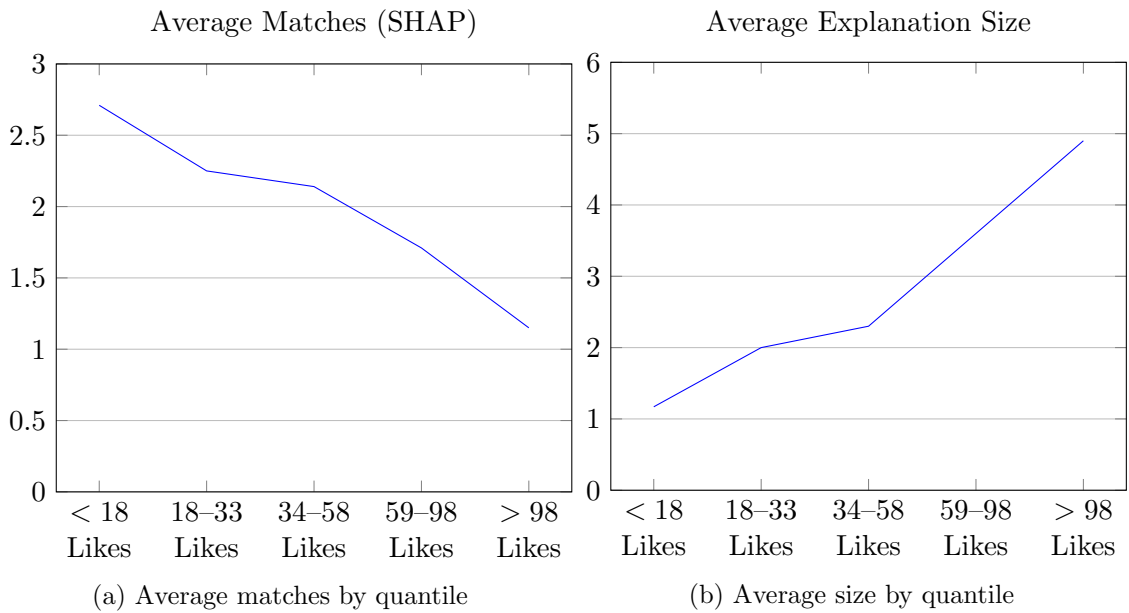

The result of the experiment is in Figure~\ref{fig:matches}, which shows the average number of matches by quantile. As predicted, SHAP approximations are not consistent for users who have Liked many pages. Recall that SHAP is supposed to be estimating the Shapley values for the features; thus, they ought to be consistent. However, for the largest instances, most cases have only one page that appears in all five SHAP runs. This implies that (for most users) the default SHAP implementation is not reliable enough to explain decisions by showing the top most-important pages.

Another alternative is to use counterfactual explanations to explain the targeting decisions, but we may worry about providing unnecessarily large explanations. We ran our algorithm to find one counterfactual explanation for each of the 500 targeting decisions. This took about 15 seconds, whereas conducting the SHAP experiment detailed above took about an hour on the same machine. The results are shown in Figure~\ref{fig:sizes}, which shows the average size of counterfactual explanations by quantile.\footnote{Recall that targeting decisions may have several counterfactual explanations. We report the average sizes of the first explanation found for each targeting decision.} The figure shows that explanations are larger for users who Liked many pages but remain relatively small considering the number of features present, which concurs with the findings of~\cite{chen2017enhancing}.


Finally, in this case study, we also adjust our method to incorporate domain-specific preferences (``costs'') and showcase how they can lead to more comprehensible explanations. The explanations shown so far (in both case studies) were generated using the heuristic search procedure proposed by~\cite{martens2014explaining}, which does not consider the relevance of the various possible explanations and was designed to find the smallest explanations first. Nonetheless, short explanations may include Likes of relatively uncommon pages, which may be unfamiliar to the person analyzing the explanation. To illustrate how domain preferences can be taken into account when generating explanations of decisions, let's say that for our problem, explanations with highly specific Likes are problematic for a feature-based explanation. The recipient of the explanation is much less likely to know these pages, so they would be better served with explanations using popular pages. To this end, we can adjust the heuristic search (as discussed in Section~\ref{sec:heuristic}) to find explanations that include more relevant---viz., more popular---pages by associating lower costs to their ``removal'' from an instance's input data. Specifically, we adjust the heuristic search to penalize less-popular pages (those with fewer total Likes) by assigning them a higher cost.

\begin{table}
    \resizebox{\textwidth}{!}{
    \begin{tabular}{c|l|l}
    \toprule
         \multirow{2}{*}{\textbf{User ID}} & \multicolumn{1}{c|}{ \multirow{2}{*}{\textbf{\shortstack{First explanation found\\ (WITHOUT the relevance heuristic)}}}} & \multicolumn{1}{c}{\multirow{2}{*}{\textbf{\shortstack{First explanation found\\ (WITH the relevance heuristic)}}}}    \\
         && \\
         \midrule
         11& `It's a Wonderful Life' (1,181 Likes)& `Reading' (47,288 Likes) \\
         & \multirow{2}{*}{\shortstack[l]{`JESUS IS LORD!!!!!!!!!!!!!!!!!!!!!!!!!!!  if you \\ know this is true press like. :)' (1,291 Likes)}} & `American Idol' (15,792 Likes) \\
         && `Classical' (8,632 Likes) \\
         \hline
         
         38& `The Hollywood Gossip' (1,353 Likes)& `Pink Floyd' (43,045 Likes) \\
         & \multirow{2}{*}{\shortstack[l]{`Remember those who have passed. Press \\ Like if you've lost a loved one' (2,248 Likes)}} & `Dancing With The Stars' (5,379 Likes) \\
         && `The Ellen DeGeneres Show' (16,944 Likes) \\
         && `American Idol' (15,792 Likes) \\
         \hline
         
         108& \multirow{2}{*}{\shortstack[l]{`Six Degrees Of Separation - The \\ Experiment' (3,373 Likes)}} & `Star Trek' (11,683 Likes) \\
         && \multirow{2}{*}{\shortstack[l]{`Turn Facebook Pink For 1 Week For \\ Breast Cancer Awareness' (12,942 Likes)}} \\
         & \multirow{2}{*}{\shortstack[l]{`They're, Their, and There have 3 distinct \\ meanings. Learn Them.' (3,842 Likes)}} & \\
         && \\
         
         \bottomrule
    \end{tabular}}
    \caption{First counterfactual explanations found}
    \label{tab:heuristic}
\end{table}

Table~\ref{tab:heuristic} shows examples of how the first explanation found by the algorithm changes depending on whether the relevance heuristic is used. As expected, the explanations found when using the relevance heuristic can include more pages than the ``shortest first'' search; however, those pages are also more popular (as evidenced by their number of Likes). Importantly, these examples show how the search procedure can be easily adapted to find context-specific explanations. In this case, the user may be interested in finding explanations with popular pages. However, the search could also be adjusted to first show the explanations with pages recently Liked by the user or pages related to the advertised product. 

\subsection{Study 3: System Decisions with Multiple Models}

Our third case study illustrates the advantages of our proposed approach when applied to complex systems, including those that use multiple models to make decisions. We use the data set from the KDD Cup 1998, available at the UCI Machine Learning Repository. The data set was originally provided by a national veteran’s organization that wanted to maximize the profits of a direct mailing campaign requesting donations. Therefore, the business problem consisted of deciding which households to target with direct mail. Importantly, one could approach this problem in several ways, such as:
\begin{enumerate}
    \item Using a regression model to predict the amount that a potential target will donate so that we can target her if that amount is larger than the break-even point.
    \item Using a classification model to predict whether a potential target will donate more than the break-even point so that we can target her.
    \item Using a classification model to predict the probability that a potential target will donate and a regression model to predict the amount if the potential target were to donate. By multiplying the results of these two models together, one could obtain the expected donation amount and send a direct mail if the expected donation is larger than the break-even point. 
\end{enumerate}

To showcase system decisions that incorporate multiple models, we illustrate our generalized framework using the third approach,  used by the winners of the KDD Cup 1998. 

We use XGBoost for both regression and classification training with 70\% of the data and the following feature subset: Age of Household Head (AGE), Wealth Rating (WEALTH2), Mail Order Response (HIT),  Male Active in the Military (MALEMILI), Male Veteran (MALEVET), Vietnam Veteran (VIETVETS), World War Two Veteran (WWIIVETS), Employed by Local Government (LOCALGOV), Employed by State Government (STATEGOV), Employed by Federal Government (FEDGOV), Percent Japanese (ETH7), Percent Korean (ETH10), Percent Vietnamese (ETH11), Percent Adult in Active Military Service (AFC1), Percent Male in Active Military Service (AFC2), Percent Female in Active Military Service (AFC3), Percent Adult Veteran Age 16+ (AFC4), Percent Male Veteran Age 16+ (AFC5), Percent Female Veteran Age 16+ (AFC6), Percent Vietnam Veteran Age 16+ (VC1), Percent Korean Veteran Age 16+ (VC2), Percent WW2 Veteran Age 16+ (VC3), Percent Veteran Serving After May 1975 Only (VC4), Number of Promotions Received in the Last 12 Months (NUMPRM12),  Number of Lifetime Gifts to Card Promotions to Date (CARDGIFT), Number of Months Between First and Second Gift (TIMELAG), Average Dollar Amount of Gifts to Date (AVGGIFT), and Dollar Amount of Most Recent Gift (LASTGIFT).

Consider the following setting. The decision-making system uses the classification and regression models on the holdout 30\% of data to target the 5\% of households with the largest (estimated) expected donations, essentially targeting the most profitable households given a limited budget. In this case, both the targeters and the targeted may be interested in explanations for why the system decided to send any particular direct mail. This is a particularly challenging problem for methods designed to explain model predictions (not decisions) because the system makes decisions using more than one model. Therefore, it is possible that the most important features for predicting the probability of donation are not the same as the most important features for predicting the donation amount, and so determining which features led to the targeting decision is not straightforward.

\begin{figure}
\begin{subfigure}{.49\textwidth}
    \centering
    \includegraphics[width=1\textwidth]{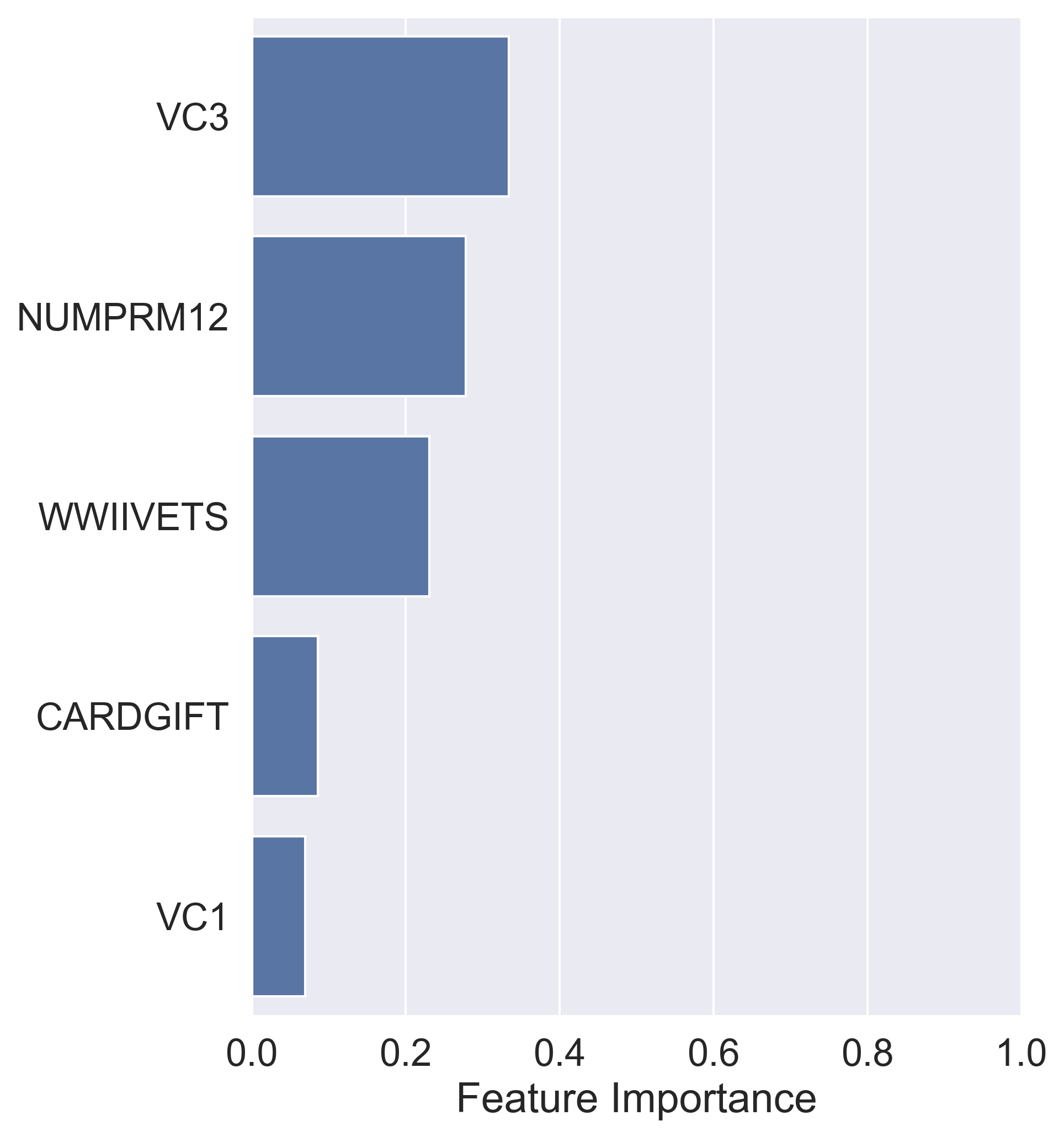}
    \caption{Top features for probability}
    \label{fig:probability}
\end{subfigure}
\begin{subfigure}{.49\textwidth}
    \centering
    \includegraphics[width=1\textwidth]{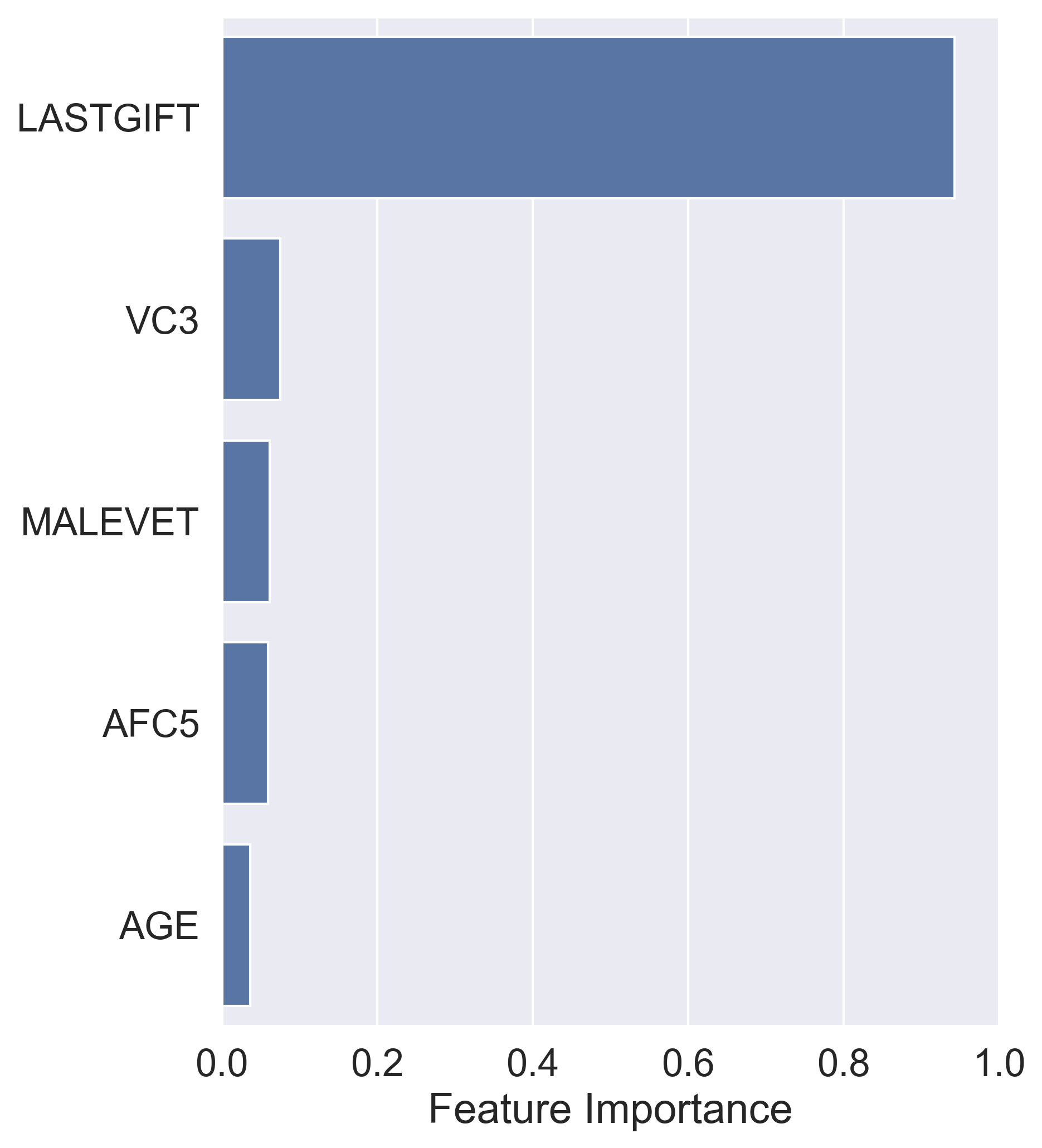}
    \caption{Top features for amount}
    \label{fig:regression}
\end{subfigure}
    \caption{Features with largest importance weights}
    \label{fig:importance}
\end{figure}

To illustrate this better, consider one targeted household in the holdout data, for which we computed SHAP values for its predicted probability of donating (given by the classification model) and its predicted donation amount (given by the regression model). We normalized the SHAP values for each model prediction so that the sum of the values adds to 1. The top five most important features for the probability prediction and the regression prediction are shown in Figure~\ref{fig:probability} and Figure~\ref{fig:regression}, respectively. Interestingly, only VC3 (percent of WW2 veterans in the household) is one of the most important features for both the classification and the regression models. We cannot explain the targeting decision from these figures alone: even though we know the most important features for each prediction, there is no way of telling what was actually vital for the system to make the targeting decision. Was the household targeted because of the size of the last gift (LASTGIFT)? Or would the household’s high probability of donating justify the targeting decision even if LASTGIFT had a smaller value?

As discussed earlier, SHAP may be used to compute feature importance weights for system decisions that incorporate multiple models  by transforming the system output into a scoring function that returns 1 if the household is targeted and 0 otherwise. However, acquiring feature importance weights for decisions made based on expected donations (rather than amounts or probabilities) would still not explain the system decisions. In contrast, counterfactual explanations can transparently be applied to system decisions that involve more than one model. Specifically, because the predicted expected donation is a scoring function (which is the result of multiplying the predictions of the two models), we can use the same framework and procedures showcased in the previous examples to find explanations for targeting decisions. Table~\ref{tab:targeting} shows the explanations found for the targeted household discussed above. 

\newcolumntype{C}{>{\centering\arraybackslash}p{3em}}

\begin{table}[ht!]
    \centering
    \begin{tabular}{c|C|C|C|C|C|C}
    \toprule
    \multirow{2}{*}{\textbf{Features}}
         & \multicolumn{6}{c}{\textbf{Explanations}} \\
    \hhline{~------}
    &\textbf{1}&\textbf{2}&\textbf{3}&\textbf{4}&\textbf{5}&\textbf{6} \\
    \midrule
    AGE &&&&&&$\downarrow$ \\
    \hline
    WWIIVETS &$\uparrow$&&&&& \\
    \hline
    VC1 &&& $\downarrow$&&& \\
    \hline
    VC2 &&&&&$\uparrow$& \\
    \hline
    VC3 &&$\uparrow$&&&&\\
    \hline
    NUMPRM12 &&$\uparrow$&$\uparrow$&$\uparrow$&$\uparrow$&$\uparrow$ \\
    \hline
    CARDGIFT &&&&$\uparrow$&& \\
    \hline
    AVGGIFT &$\uparrow$&$\uparrow$&$\uparrow$&$\uparrow$&$\uparrow$&$\uparrow$ \\
    \hline
    LASTGIFT &$\uparrow$&$\uparrow$&$\uparrow$&$\uparrow$&$\uparrow$&$\uparrow$ \\
    \midrule
    \multicolumn{7}{l}{$\uparrow$ means household was targeted because feature is \textbf{above} average. }\\
    \multicolumn{7}{l}{$\downarrow$ means household was targeted because feature is \textbf{below} average.} \\
    \bottomrule
    \end{tabular}
    \caption{Explanations for targeting decision}
    \label{tab:targeting}
\end{table}

Interestingly, some of the highest-scoring SHAP features, shown in Figure~\ref{fig:importance},  are not present in any of the explanations (e.g., MALEVET), whereas some features that are present in some explanations do not have large SHAP values (e.g., AVGGIFT). In fact, AVGGIFT had a negative SHAP value in the regression model (meaning we would expect its impact on the non-default decision to be negative), but it appears in all explanations. This example illustrates the importance of defining explanations in terms of decisions and not predictions, particularly when dealing with complex, non-linear models, such as XGBoost. 

More specifically, because SHAP attempts to evaluate the overall impact of features on the model prediction, it averages out the negative and positive impacts that features have on the prediction when their values are changed alongside all other feature combinations. Hence, if a feature has a large negative impact in one feature ordering and several small positive impacts in other orderings, that feature may have a negative SHAP value if the single negative impact is greater than the sum of the small positive impacts. This behavior is the same as illustrated by Example 3 in Section~\ref{sec:example3}, which, of course, would be counterproductive when trying to understand the influence of features on decision making. Averaging the impact of features over all feature orderings hides the fact that, in non-linear models, features may provide evidence in favor or against a decision depending on what other features are changed, which explains why AVGGIFT had a negative SHAP value but is present in the explanations shown in Table~\ref{tab:targeting}. 

\section{Discussion}

These studies with real-world data illustrate various advantages of counterfactual explanations over importance weighting methods. The first study shows that the importance weights of features are not enough to determine how they affect system decisions. It also shows how different imputation methods may be used to generate and customize counterfactual explanations for various purposes and users. The second study demonstrates the strengths of counterfactual explanations in the presence of high-dimensional data. In particular, the study shows that sampling-based approximations of importance weights become worse as the number of features increases. Counterfactual explanations sidestep this issue because small subsets of features are usually enough to explain decisions. Moreover, the study showcased a heuristic procedure to search for and sort counterfactual explanations according to their relevance. Finally, the third study shows that importance weights may be misleading when decisions are made using multiple (and complex) models. More specifically, we see a real instance of the phenomenon in the synthetic example in Section~\ref{sec:example3}, in which features with non-positive SHAP weights may have a positive effect on system decisions. 

It has been argued that a disadvantage of counterfactual explanations is that each instance (decision) usually has multiple explanations~\citep{explanationssite}; this is also referred to as the Rashomon effect. The argument is that multiple explanations are inconvenient because people may prefer simple explanations over the complexity of the real world.\footnote{Of course, this "disadvantage" could be avoided simply by showing only one of the counterfactual explanations, which could be a unique solution by defining a preference order over feature subsets, as we have discussed.} This issue may be exacerbated as the number of features increases because the number of counterfactual explanations may grow exponentially. In contrast, most importance weighting methods converge to a unique solution in theory (e.g., Shapley values in the case of SHAP), if allowed enough run time. 

However, as we have shown, instance-weight explanations have serious deficiencies for explaining decisions, so we cannot argue that they are preferable because they are simpler. Moreover, our second case study shows that importance weighting methods may not scale well when the number of features increases because their approximations may become inconsistent. In the case of counterfactual explanations, measures of relevance (e.g., number of Likes in our Facebook case study) may be incorporated as part of the heuristic procedures used to find and rank counterfactual explanations. Thus, the fact that there are multiple counterfactual explanations is not necessarily problematic. Our study could find short, consistent, and relevant explanations significantly faster than it could compute importance weights, even with many features. 

A factor briefly explored in the first case study is the sensitivity of the explanations to the method used to produce counterfactual values. This is not a challenge particular to the counterfactual approach; feature importance approaches, such as SHAP, also require the choice of such a method (e.g., mean imputation). We argue that the evidence-based perspective is useful to define plausible and relevant counterfactual scenarios, and so the choice of the method should carefully match the domain and the intent behind the explanations. This is an interesting direction for future research, as we would expect distinct alternatives for dealing with evidence removal to affect explanations differently, resulting in different interpretations of the system decision. For example, mean and model-based imputation can produce very different explanations, and each may be more appropriate in different settings (as shown in our first case study). 

Another option is to forgo imputation strategies altogether and produce counterfactual values using an alternative procedure. For example, continuous features could be discretized into bins, and the algorithm we propose could be easily adapted to greedily move feature values across bins until the predicted class is changed, as done by~\cite{gomez2020vice}. This procedure could be guided by the preference function introduced in our framework to search first for the most relevant counterfactual values. Similarly, any of the several methods proposed by~\cite{lash2017generalized} for inverse classification could also be used to produce counterfactual explanations. 

Importantly, our study compares importance weights with a specific type of counterfactual explanation. As defined in Section~\ref{sec:definition}, our explanations use the evidence-based perspective to simulate counterfactual worlds in which the evidence supporting the decision made by the system is absent. Nonetheless, other types of counterfactual worlds may be of interest when explaining decisions. For example, our first case study showed that some loan applicants were denied credit because the amount they requested was larger than average. While this explains the credit denial decision, these applicants may instead be interested in the maximum amount they could be approved for. Such a counterfactual explanation could be defined as a set of ``minimal'' feature adjustments that changes the decision.

Other researchers have proposed various methods to obtain such counterfactual explanations~\citep{verma2020counterfactual}. For example, in the context of explaining predictions (not decisions), \cite{wachter2017counterfactual} define counterfactual explanations as the smallest change to feature values that changes the prediction to a predefined output. Thus, they address explanations as a minimization problem in which larger (user-defined) distances between counterfactual instances and the original instance are penalized more. However, their method (i) focuses on models for which the gradient at the decision point can be computed, (ii) does not work with categorical features, and (iii) may require access to the machine learning method used to learn the model (which usually is not available for deployed decision-making systems). \cite{tolomei2017interpretable} define counterfactual explanations similarly but instead propose how to find such explanations when using tree-based methods. Other counterfactual methods have been implemented in the Python package Alibi.\footnote{See \url{https://github.com/SeldonIO/alibi}} The package includes a simple counterfactual method loosely based on~\cite{wachter2017counterfactual} and an extended method that uses class prototypes to improve the interpretability and convergence of the algorithm~\citep{van2019interpretable}.

Another key assumption behind all the instance-level explanation methods discussed in this paper (feature importance and counterfactual) is that examining an instance's features will make sense to the user. This presumes that the features themselves are comprehensible, which would not be the case, for example, if the features are too low level or for cases where the features have been obfuscated, for example, to address privacy concerns (see, e.g., the discussion of ``doubly deidentified data'' by~\cite{provost2009audience}). 

Another promising direction for future research is studying how users perceive these different sorts of explanations. It would be interesting to analyze the impact that various types of explanations have on users' adoption and interpretation of AI systems, preferably through user studies~\citep{binns2018s,dodge2019explaining}. \cite{kaur2019interpreting} studied data scientists' use of interpretability tools (including SHAP) when uncovering common issues that arise when building and evaluating predictive models. They found that despite being provided with standard tutorials, few participants could accurately describe what the visualizations were showing. As a result, some participants over-trusted the model because they used the explanations to rationalize suspicious observations, whereas others became skeptical of the visualizations and eventually stopped using them.

Importantly, different users are likely to require different information from explanations, and thus it is unlikely that a particular explanation will always be the best for every objective. Other researchers have recognized this and have proposed general frameworks to define characteristics of good explanations based on user needs~\citep{lu2019good}. This study builds on this school of thought by providing a flexible explanation framework that may be used to address a wide and diverse range of needs. We discuss this in more detail below as part of the managerial implications of our study.

Additionally, how explanations are visualized is also likely to affect how users perceive them. Although we used text and tables to present counterfactual explanations in this paper, interactive tools are probably much better suited for visualizing counterfactual explanations, especially if the goal is to analyze decisions at an aggregate level. What would work best in practice to visualize explanations is also likely to be context dependent, but there is already a nascent stream of research proposing tools to visualize counterfactual explanations as proposed by~\cite{martens2014explaining}. Examples include ~\cite{tamagnini2017interpreting,krause2017workflow,gomez2020vice}. Any of these visualization tools could be easily adapted to display the counterfactual explanations proposed by our framework. 

Another interesting direction is to learn from data the explanations that work better for different users. In the context of consumer-facing recommendations, \cite{mcinerney2018explore} propose a method that simultaneously learns the best content to recommend for each
user and the type of explanation(s) to which each user responds best. They showcase their method with music recommendations and find that personalizing explanations and recommendations together significantly increases user engagement. Therefore, explanations may also have an important role in the outcome that the system decisions seek to optimize. This type of method could be used in conjunction with our framework to learn the imputation strategies or cost functions that are most effective to improve decisions (or other outcomes).  

Having said this, more work is needed to provide explanations that truly address user needs and, in particular, the needs of decision subjects rather than decision makers. \cite{barocas2020hidden} reveal several easily overlooked assumptions on which uses of counterfactual explanations  (and explanation methods in general) often rely, and that may have negative effects on users of system decisions, specifically, when those explanations are then used to suggest concrete actions. For example, feature changes may interact with facts about a person's life that are invisible to the model, and thus the system may recommend something that would interfere with another goal in the person's life. Changes that might be inexpensive for one person might be costly for another person, and thus we may need to account for how feature changes affect costs and how costs vary across the population. In other cases, highlighting what the person must \emph{not} change might be as important as the explanation itself: think about a mortgage applicant who switched jobs to increase her annual income only to find out that she now does not qualify for a home loan due to her short time of employment at her new job.   

While our framework could be used to work around some of these problems (e.g., decision subjects could communicate their costs by ranking features according to how easy they would be to change), these are unresolved challenges that require careful solutions to avoid other potential issues (e.g., privacy concerns, excessive complexity, users gaming the system). To address these challenges, future research should seek to understand what actions people actually take when confronted with explanations and how they are affected by those actions.

\section{Managerial Implications}

Importance weighting methods are rapidly becoming a popular (if not the most popular) alternative for explaining model predictions. However, this paper shows that these methods may not be appropriate to explain the decisions made by model-based AI systems. Notably, the examples and case studies in this paper illustrate various pitfalls that managers should be aware of when deploying importance-weight explanations. The most salient is that importance weights are insufficient to communicate whether and how features affect decisions. Our paper proposes a counterfactual framework as an alternative specifically designed for such a task, illustrating its advantages throughout the examples.

We also demonstrate how counterfactual explanations can be applied much more broadly to more problems and systems---than many prior authors seem to have realized.  Our case studies use the proposed counterfactual explanations to explain system decisions made (a) using numeric and categorical features, (b) in low- and high-dimensional settings, (c) with linear and non-linear models, and (d) with system decisions based on one or multiple predictive models. The examples and case studies were also motivated by various business settings commonly encountered in practice and in which AI systems may be particularly useful, such as credit scoring and targeted advertising.

Finally, we propose two ways managers or other end users may tailor counterfactual explanations to suit their context. The first consists of defining how to deal with the removal of evidence to generate explanations. This choice should be driven by business context and may change according to stakeholder needs. Continuing with the targeted advertising example, mean and mode imputation may be a reasonable approach for users who want to understand which of their actions led the system to target them. At the same time, a manager using the system to make those targeting decisions may want to understand which features led to the best targeting decisions to decide which features to keep investing in (assuming the manager is using purchased third-party data for targeting, which is not unusual). Thus, for the manager’s use case, a better approach to deal with evidence removal might be to simulate the system behavior if the manager were to stop purchasing data for that feature, for example, by using a model built without those features. 

The second consists of allowing end-users to tailor explanations by incorporating context information as part of the heuristic procedure used to generate counterfactual explanations. Such information could consist of the cost of acquiring or changing the features, the degree of relevance of the features, or other domain-driven ``rules'' ( e.g., the feature value for ``is\_female'' should not be changed if ``is\_pregnant=1'').  We propose (and illustrate how) to incorporate this information as part of a user-defined cost function that a heuristic procedure may use to search first for potential explanations with ``lower costs,'' resulting in more context-specific explanations.

\section{Conclusion}

We examine the problem of explaining data-driven decisions made by AI systems from a causal perspective: if the question we seek to answer is ``why did the system make a specific decision,” we can ask ``which inputs caused the system to make its decision?” This approach is advantageous because (a) it standardizes the form that an explanation can take; (b) it does not require all features to be part of the explanation, and (c) the explanations can be separated from the specifics of the model. Thus, we define a (counterfactual) explanation as a set of features that is causal (meaning that changing the values of these features changes the decision) and irreducible (meaning that changing the values of any subset of the features in the explanation does not change the decision).

Importantly, this paper shows that explaining model predictions is not the same as explaining system decisions because features that have a large impact on predictions may not have an important influence on decisions. Moreover, we show through examples and case studies that the increasingly popular approach of explaining model predictions using importance weights has significant drawbacks when repurposed to explain system decisions. In particular, we demonstrate that importance weights may be ambiguous or even misleading when the goal is to understand how features affect a specific decision.

Our work generalizes previous work on counterfactual explanations in two important ways: (i) we explain system decisions (which may incorporate predictions from several predictive models using features with arbitrary data types) rather than model predictions, and (ii) we do not enforce any specific method to produce counterfactual values for the features. Finally, we also propose a heuristic procedure that allows the tailoring of explanations to domain needs by introducing costs, for example, the costs of changing the features responsible for the decision.  

\bibliography{bibliography}

\end{document}